\documentclass[conference]{IEEEtran}
\ifCLASSINFOpdf
\else
\fi

\usepackage[T1]{fontenc}
\usepackage[utf8x]{inputenc}
\usepackage{float}
\usepackage{graphicx}
\usepackage{subfigure}
\usepackage[numbers, sort&compress]{natbib}
\usepackage{amssymb}
\usepackage{bm}
\usepackage{amsmath}
\usepackage{mathrsfs}

\floatstyle{ruled}
\newfloat{algorithm}{tbp}{loa}
\floatname{algorithm}{Algorithm}

\begin{document}
%
% paper title
% Titles are generally capitalized except for words such as a, an, and, as,
% at, but, by, for, in, nor, of, on, or, the, to and up, which are usually
% not capitalized unless they are the first or last word of the title.
% Linebreaks \\ can be used within to get better formatting as desired.
% Do not put math or special symbols in the title.
\title{Possibilistic Fuzzy Local Information C-Means for Sonar Image Segmentation}

% author names and affiliations
% use a multiple column layout for up to three different
% affiliations
\author{\IEEEauthorblockN{Alina Zare, Nicholas Young, Daniel Suen}
\IEEEauthorblockA{Electrical and Computer Engineering\\
University of Florida\\
Gainesville, FL 32611\\
\{azare, ndy14, dsuen1\}@ufl.edu}
\and
\IEEEauthorblockN{Thomas Nabelek, Aquila Galusha, James Keller}
\IEEEauthorblockA{Computer Science and Electrical Engineering\\
University of Missouri\\
Columbia, MO 65211\\
kellerj@missouri.edu}}

% make the title area
\maketitle

% As a general rule, do not put math, special symbols or citations
% in the abstract
\begin{abstract}
Side-look synthetic aperture sonar (SAS) can produce very high quality images of the sea-floor.  When viewing this imagery, a human observer can often easily identify various sea-floor textures such as sand ripple, hard-packed sand, sea grass and rock.  In this paper, we present the Possibilistic Fuzzy Local Information C-Means (PFLICM) approach to segment SAS imagery into sea-floor regions that exhibit these various natural textures.  The proposed PFLICM method incorporates fuzzy and possibilistic clustering methods and leverages (local) spatial information to perform soft segmentation.  Results are shown on several SAS scenes and compared to alternative segmentation approaches. 
\end{abstract}

% no keywords

% For peer review papers, you can put extra information on the cover
% page as needed:
% \ifCLASSOPTIONpeerreview
% \begin{center} \bfseries EDICS Category: 3-BBND \end{center}
% \fi
%
% For peerreview papers, this IEEEtran command inserts a page break and
% creates the second title. It will be ignored for other modes.
\IEEEpeerreviewmaketitle

\section{Introduction}
% no \IEEEPARstart
High-resolution Synthetic Aperture SONAR (SAS) systems have the image resolution and detail needed to be able to distinguish between different sea-floor environmental contexts such as sand ripple, hard-packed sand and sea grass. These systems also have the resolution and detail needed for automated target object detection and classification.  However, underwater target objects may display varying characteristics across environmental contexts.   For example, target signatures may be hidden or masked to varying degrees depending on the sea-floor type in which they are found. Therefore, methods that can identify the seabed environment can be used to within an automated scene understanding system or to assist in target classification and detection in an environmentally adaptive or context-dependent approach.  Furthermore, the boundaries between sea-floor types are often gradual with wide regions of transition \cite{Cobb:2014}. Thus, a soft segmentation approach that can identify and characterize these gradual transition regions are necessary.  

In this paper, we present a soft segmentation algorithm that leverages local spatial information to encourage smooth segmentation maps and estimates both possiblistic typicality and fuzzy membership  values to  identify outliers and to discriminate between the various sea-floor types.

\section{Proposed Method: PFLICM}
Our proposed PFLICM algorithm merges previous approaches for possibilistic fuzzy clustering and soft segmentation of imagery.  Namely, PFLICM combines the Possibilistic-Fuzzy Clustering algorithm, PFCM \cite{Pal2005}, and the Fuzzy Local Information C-Means, FLICM, algorithm \cite{Krinidis2010}. These approaches are blended by PFLICM through the objective function shown in \eqref{eqn:fobj}:

\begin{eqnarray}
J &=&  \sum_{c=1}^{C}\sum_{n=1}^{N}au_{cn}^{m}   \left( \left\|\mathbf{x}_{n}-\mathbf{c}_{c}\right\|^2_2 + G_{cn}\right)\nonumber \\
& &+bt_{cn}^q   \left\|\mathbf{x}_{n}-\mathbf{c}_{c}\right\|^2_2 +  \sum_{c=1}^{C}\gamma_c \sum_{n=1}^{N} (1 - t_{cn})^q
  \label{eqn:fobj}
\end{eqnarray}
such that 
\begin{eqnarray}
& u_{cn}\geq 0 \quad \forall n = 1, \ldots, N; \quad \sum_{c=1}^C u_{cn} = 1
\label{eqn:PropMemConstraints}
\end{eqnarray}
where $\mathbf{x}_{n}$ is a $d\times 1$ column vector representing the $n^{th}$ pixel, $C$ is the number of clusters being estimated, $\mathbf{c}_{c}$ is a $d\times 1$ vector of $c^{th}$ cluster center, weight $u_{cn}$  is the membership value of the $n^{th}$ pixel in the $c^{th}$ cluster, $t_{cn}$ is the typicality value of the $n^{th}$ pixel in the $c^{th}$ cluster, $a, b$, and $\alpha$ are fixed parameter values used to balance the terms of the objective function, and $m$ and $q$ are fixed ``fuzzifier'' parameters which control the degree of sharing across the clusters and the degree to which points may be labeled as outliers, respectively. Also, 
\begin{equation}
G_{cn} = \sum_{\substack{k \in \mathscr{N}_n \\ k \ne n}} \frac{1}{d_{nk} + 1}(1 - u_{ck})^m \left|\left| \mathbf{x}_k - \mathbf{c}_{c} \right|\right|_2^2,
\label{eqn:fuzzyG}
\end{equation}
where $\mathbf{x}_n$ is the center pixel in the local window under consideration, $\mathscr{N}_n$ is the neighborhood around the center pixel (i.e., a neighborhood defined by fixed radius), $d_{nk}$ is the Euclidean distance between the image indices between $\mathbf{x}_n$ and $\mathbf{x}_k$.  These terms are further defined and explained in the following subsections. 

\subsubsection{Soft Segmentation with Local Information} The first term in \eqref{eqn:fobj},  $\sum_{c=1}^{C}\sum_{n=1}^{N}au_{cn}^{m}   \left( \left\|\mathbf{x}_{n}-\mathbf{c}_{c}\right\|^2_2 + G_{cn}\right)$, follows directly from the objective function of the FLICM algorithm \cite{Krinidis2010}.  This term combines fuzzy clustering \cite{bezdek1984fcm} (and the estimation of fuzzy membership values) with local spatial information to encourage neighboring pixels to have similar fuzzy membership values.  In particular, during minimization of \eqref{eqn:fobj}, larger fuzzy membership values, $u_{cn}$, for a data point $n$ will be assigned to clusters in which the both the distance between the data point and the cluster center is relatively small $(\left\|\mathbf{x}_{n}-\mathbf{c}_{c}\right\|^2_2)$ and where the $G_{cn}$ value is small. The $G_{cn}$term can be interpreted as a penalty for clusters in which neighboring pixels have a small membership value.  
\subsubsection{Outlier Identification with Possiblistic Clustering}  The inclusion of the  $bt_{cn}^q \left\|\mathbf{x}_{n}-\mathbf{c}_{c}\right\|_2^2$ and $ \gamma_c (1-t_{cn})^q   $ terms provide for the identification of outliers in the input SAS imagery.  These terms are adapted from the Possibilisitc clustering \cite{Krishnapuram:1993} and Possibilistic-Fuzzy clustering algorithms \cite{Pal2005}.  Possibilistic clustering methods estimate the \emph{typicality}, $t_{cn}$, of each data point and, unlike the membership values, $u_{cn}$, are not constrained to sum-to-one across clusters. When the typicality of a data point is close to zero for a particular cluster, that indicates that the data point is an outlier with respect to that cluster.   A data point that is an outlier in all clusters (i.e., sufficiently far from all cluster centers) will have a low typicality value in all clusters.  Thus, the estimated typicality values identify the outliers in an input data set and  prevent them from influencing cluster estimation and baising the resulting cluster representative values.   This is an important feature for locating noise pixels as well as for excluding anomolous (e.g., target object) pixels while characterizing the background and sea-floor.  In this way, potential target objects in a scene do not affect the sea-floor segmentation results.

\subsubsection{Alternating Optimization Approach}
Cluster centers, membership, and typicality values are estimated using alternating optimization.  All values are initialized and, then, the parameters are updated iteratively.  Solving $ \frac{\partial J}{\partial \mathbf{c}_{c}} = 0$ results in the following update equation for the endmembers, 
\begin{align}
  \mathbf{c}_c = \frac{\sum_n(a u_{cn}^r + b t_{cn}^q) \mathbf{x}_n }{\sum_n (a u_{cn}^r + b t_{cn}^q)} 
   \label{eqn:fuzzpop-e-update}
\end{align}
The update equation for the membership values is found by adding a Lagrange multiplier term to enforce the sum-to-one constraint on the membership values, resulting in:
\begin{equation}
  u_{cn} = \frac{1}{\sum_{k=1}^C \left(
    \frac{(\mathbf{x}_n - \mathbf{c}_c)(\mathbf{x}_n - \mathbf{c}_c)^T + G_{cn}}{(\mathbf{x}_n - \mathbf{c}_k)(\mathbf{x}_n - \mathbf{c}_k)^T + G_{kn}}\right)^{\frac{1}{r-1}}}
  \label{eqn:fuzzpop-u-update}
\end{equation}

The update equation for the typicalities value is found to be:
\begin{equation}
  t_{cn} = \frac{1}{1 + \left(\frac{b}{\gamma_c} ||\mathbf{x}_n - \mathbf{c}_c ||_2^2\right)^{\frac{1}{q-1}}}
  \label{eqn:fuzzpop-t-update}
\end{equation}
where the $\gamma_c$ value is set to be the mean of the $\left|| \mathbf{x}_{n}-\mathbf{c}_{c}\right||_2^2$ value for all of the pixels in the associated cluster. Therefore, each iteration, the $\gamma_i$ value gets updated.  

 Given these update equations, the proposed PFLICM algorithm is summarized below.

\begin{algorithm}[H]
\begin{raggedright}
\caption{ PFLICM }

\par\end{raggedright}

\begin{raggedright}
Initialize membership values to $\frac{1}{C}$, typicality values to $1$, and randomly select input data points as the $C$ initial cluster centers. 
\par\end{raggedright}

\begin{description}
\item [{REPEAT}]~

Update cluster centers $\left\{ \mathbf{c}_{c}\right\} _{c=1,\ldots,C}$
using \eqref{eqn:fuzzpop-e-update}

Update the memberships using  \eqref{eqn:fuzzpop-u-update}

Update the typicalities using \eqref{eqn:fuzzpop-t-update}

\item [{UNTIL}] $\quad$convergence
\end{description}

\end{algorithm}

\section{Application to SAS Imagery}
Our application of PFLICM to SAS imagery for sea-floor segmentation is comprised of three major components: (1) feature extraction, (2) superpixel segmentation, and (3) application of PFLICM.  The first component, feature extraction, is carried out in order to produce segmentation results that fall along the boundaries of the distinct sea-floor textures.  The features extracted were selected to highlight the differences between textures of interest.  The features used were a Sobel edge histogram descriptor \cite{frigui2009detection}, histogram of oriented gradients \cite{dalal2005histograms}, and local binary pattern features \cite{guo2010completed}.   The Sobel edge histogram descriptor was generated by applying the Sobel edge detector for vertical, diagonal, horizontal and anti-diagonal edges.  Then, each pixel is labeled with the edge type that returned the largest Sobel edge detection value or it is marked as ``no edge'' if all of the detection values were below a prescribed threshold of 0.5.  Then, the feature vector associated with each pixel is the histogram of edge types for the surrounding $17 \times 17$-size neighborhood.  The HOG features were generated using cell and block sizes of $2\times2$, a 1 pixel block overlap, 9 bins, computed over a sliding window size of $5\times 5$.   Finally, the LBP features were generated using $3\times3$ sized cells.  

\begin{figure*}[!t]
	\center{
		\subfigure[ Original SAS Image ]{\resizebox{!}{2in}{\includegraphics{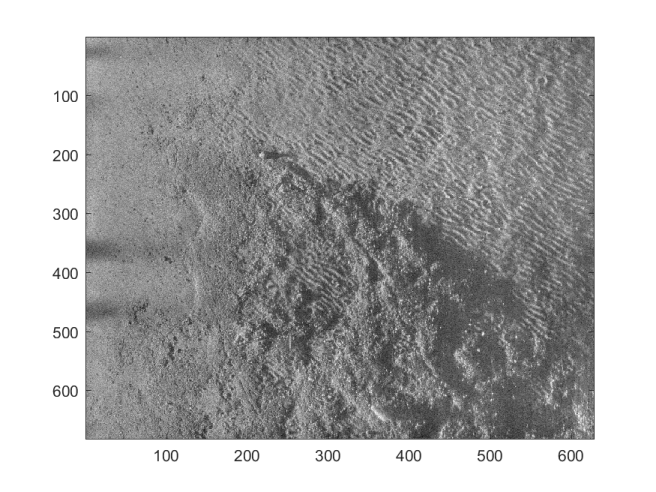}}}
		\subfigure[ Superpixel Segmentation of SAS Image ]{\resizebox{!}{2in}{\includegraphics{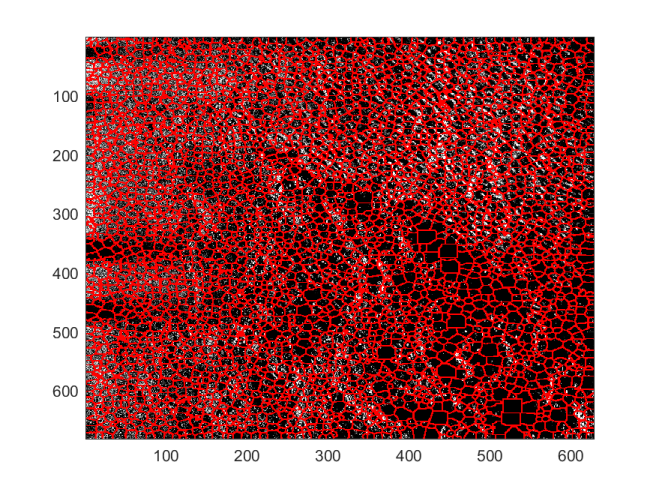}}}
     		\caption[]{ (a) Original SAS image shows a variety of textures including ``flat sand,'' ``sand ripple,'' and ``rock.'' (b) Superpixel segmentation of the image into 3500 superpixels. }
		\label{fig:seg}		}
\end{figure*}

As opposed to applying PFLICM to every individual pixel in a SAS image, in our implementation input grayscale SAS imagery are first oversegmented using the SLIC superpixel segmentation algorithm \cite{achanta2012slic}.  Then, the feature vectors associated with each resulting superpixel are averaged and center location of the superpixel is marked as the location (i.e., the image row and column) to which this average feature vector corresponds.  Then, PFLICM is applied to only these feature vectors that summarize each superpixel.  The use of superpixels (instead of individual pixels from the input image) significantly reduces the overall computation required.  Figure \ref{fig:seg} shows an example input image and its resulting superpixel segmentation.  

After feature extraction and superpixel segmentation, the PFLICM algorithm is applied to the set of average feature vectors associated with each superpixel.  After applying PFLICM, the resulting segmentation can be examined by creating the \textit{product maps} which are generated by computing the product of the estimated membership and typicality value for each superpixel and, then, assigning this value to each individual pixel associated with that superpixel. Specifically, the value of a product map, $\mathbf{P}$ at pixel $n$ for cluster $c$ is computed as $\mathbf{P}_c(n) = u_{cn}t_{cn}$.

\section{Experimental Results}

In the first set of results PFLICM was applied to seven images with the following parameter settings: 
\begin{itemize}
	\item Membership weight $a$ = 14
	\item Typicality weight $b$ = 0.3
	\item Fuzzifier for Fuzzy Clustering Term $m$ = 1.8
	\item Fuzzifier for Possibilistic Clustering Term $q$ = 2.2
	\item Change threshold (stopping criteria) $\epsilon$ = 1e-6
	\item Number of clusters = 3
	\item Radius of neighborhood window $\mathscr{N}_n$  = $\infty$ (All superpixels in the image were considered to be in the same neighborhood.)
\end{itemize}

The parameters were determined manually.  For comparison, each image was also segmented using the FLICM algorithm and the PFCM algorithm (using the same parameter settings as applied for PFLICM for the associated parameters across the algorithms). Unless otherwise noted, each set of images displays the most relevant values - that is, the product maps are shown for PFLICM and PFCM while the membership maps are shown for FLICM (since FLICM does not generate typicality values). 

The results of the seven images are shown in Figs. \ref{fig:im4} and \ref{fig:im1} through \ref{fig:im7}.
\begin{figure}[!h]
  \center{
      \resizebox{!}{2.5in}{\includegraphics[trim={1cm 0cm 1cm .75cm},clip]{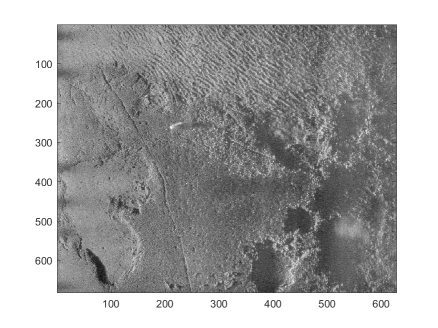}}\\
        \caption[]{ SAS image containing sand ripples, flat sand with large holes, and an outlier. }
  }
\end{figure}

\begin{figure}[!h]
	\center{
		\subfigure[ PFLICM Cluster 1 ]{\resizebox{!}{.95in}{\includegraphics[trim={1.6cm 1cm 3cm .75cm},clip]{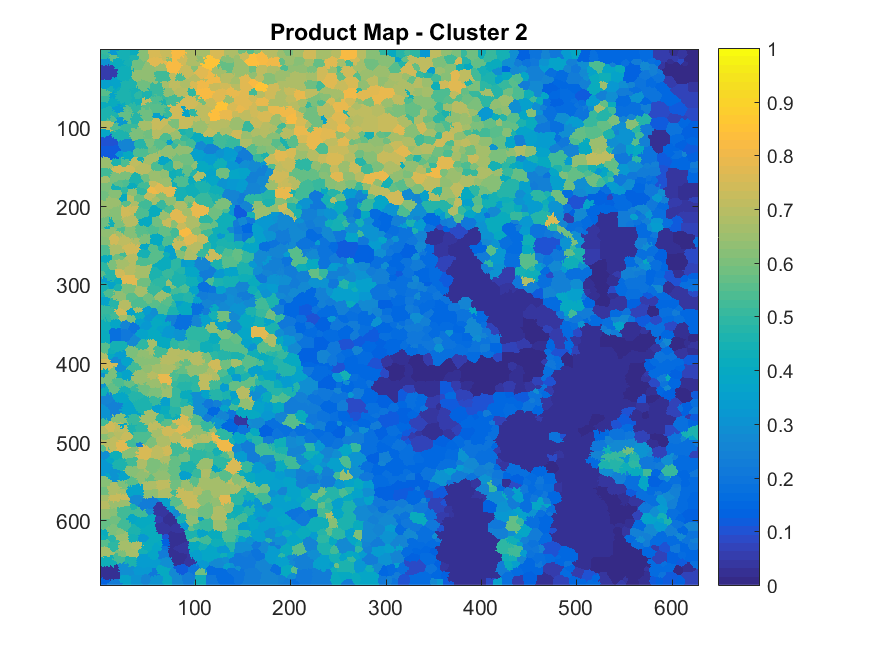}}}
		\subfigure[ PFLICM Cluster 2 ]{\resizebox{!}{.95in}{\includegraphics[trim={1.6cm 1cm 3cm .75cm},clip]{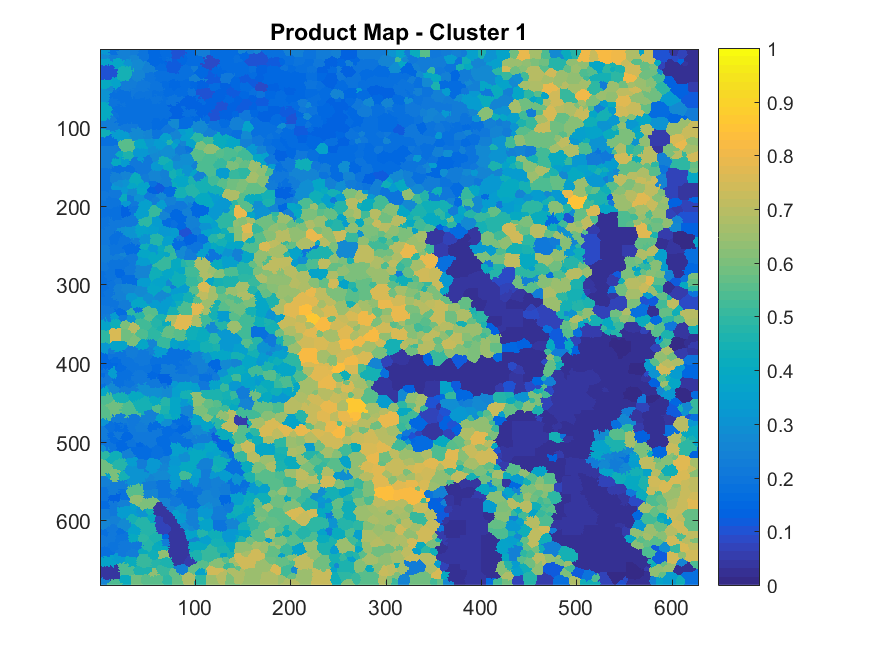}}}
		\subfigure[ PFLICM Cluster 3 ]{\resizebox{!}{.95in}{\includegraphics[trim={1.6cm 1cm 3cm .75cm},clip]{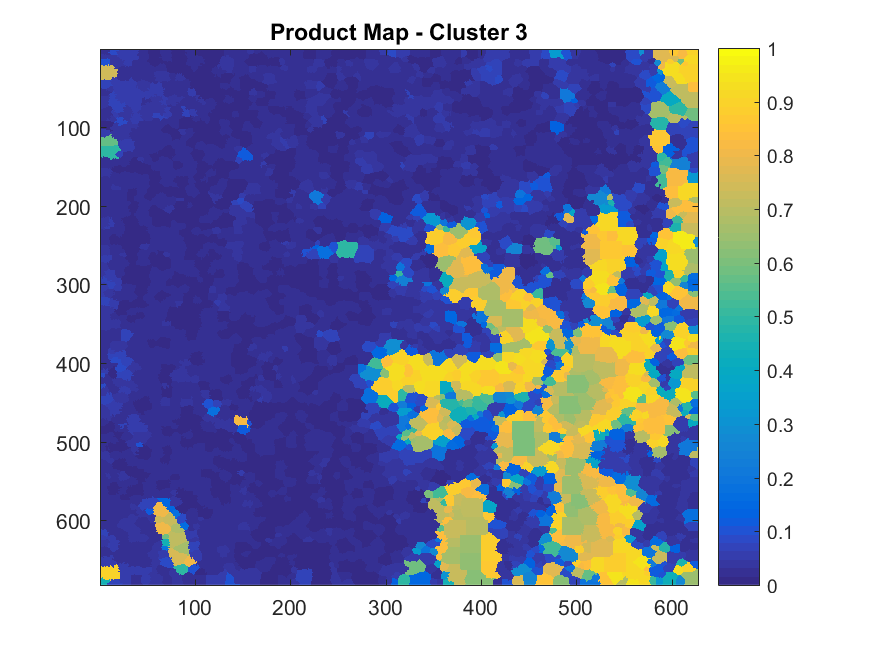}}}\\
		\subfigure[ FLICM Cluster 1 ]{\resizebox{!}{.95in}{\includegraphics[trim={1.6cm 1cm 3cm .75cm},clip]{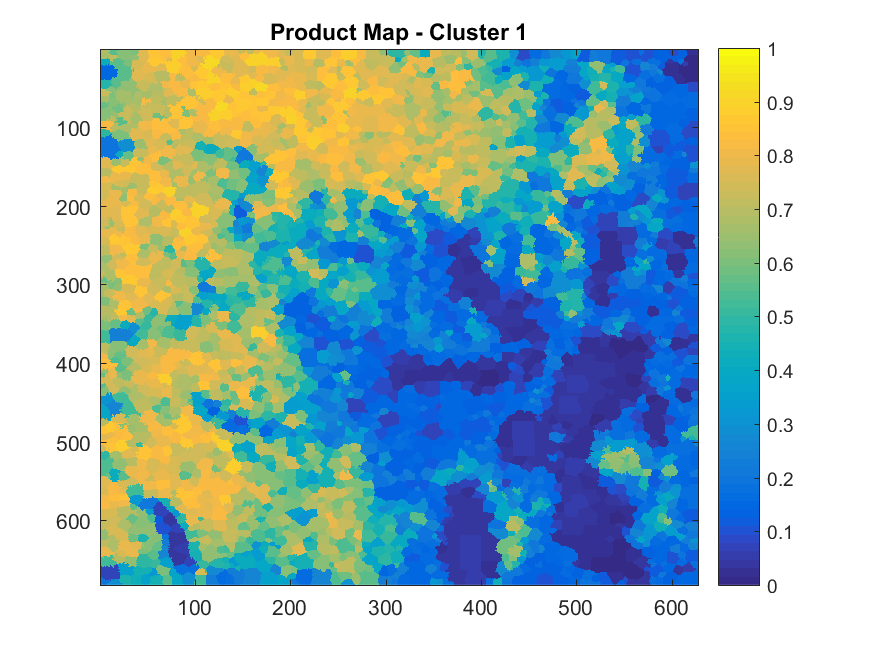}}}
		\subfigure[ FLICM Cluster 2 ]{\resizebox{!}{.95in}{\includegraphics[trim={1.6cm 1cm 3cm .75cm},clip]{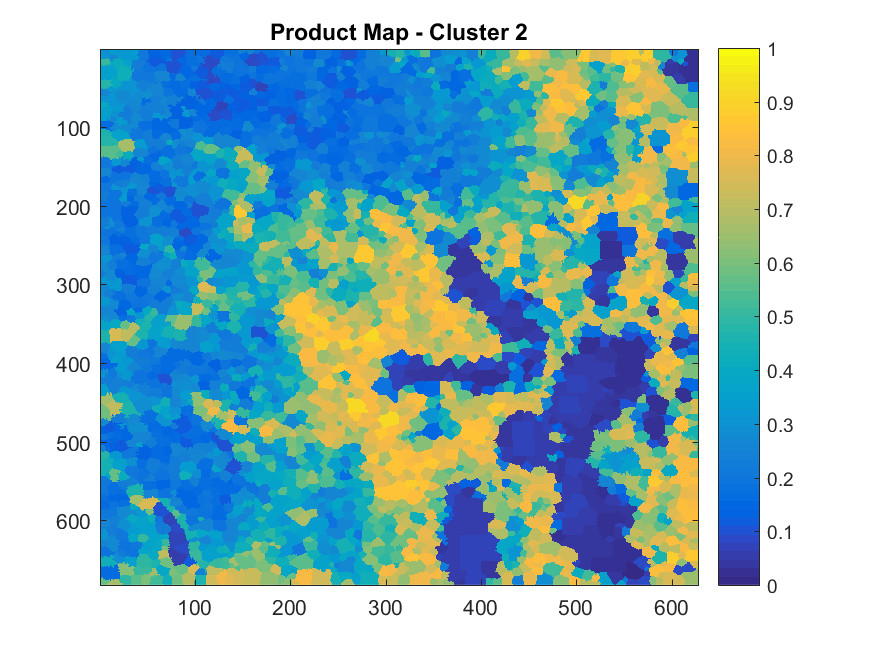}}}
		\subfigure[ FLICM Cluster 3 ]{\resizebox{!}{.95in}{\includegraphics[trim={1.6cm 1cm 3cm .75cm},clip]{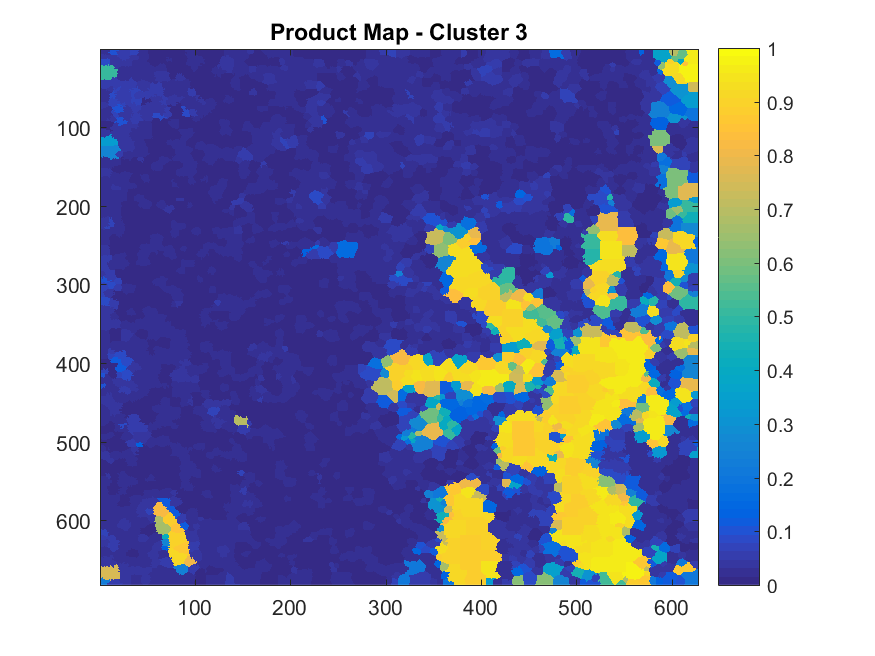}}}\\
		\subfigure[ PFCM Cluster 1 ]{\resizebox{!}{.95in}{\includegraphics[trim={1.6cm 1cm 3cm .75cm},clip]{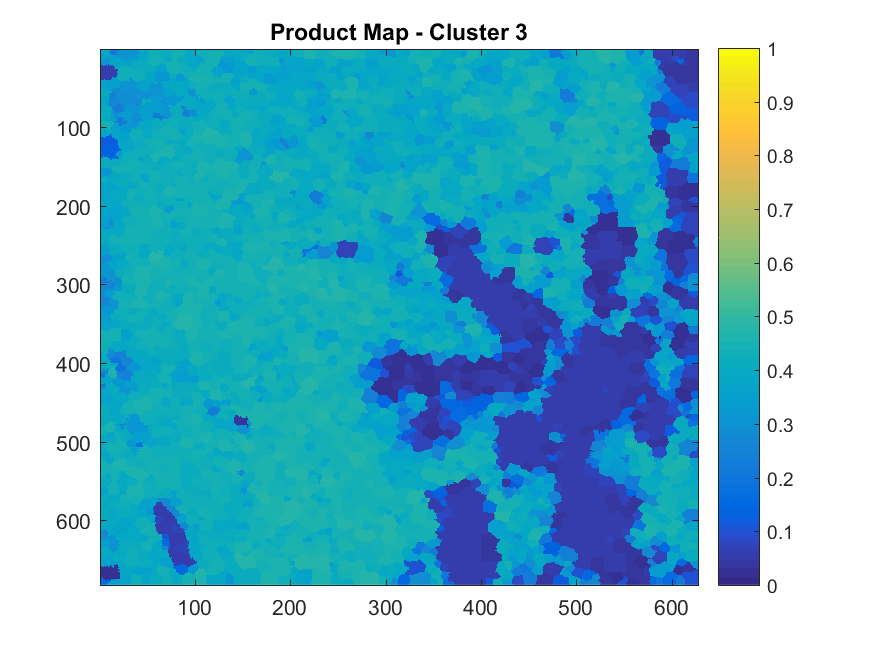}}}
		\subfigure[ PFCM Cluster 2 ]{\resizebox{!}{.95in}{\includegraphics[trim={1.6cm 1cm 3cm .75cm},clip]{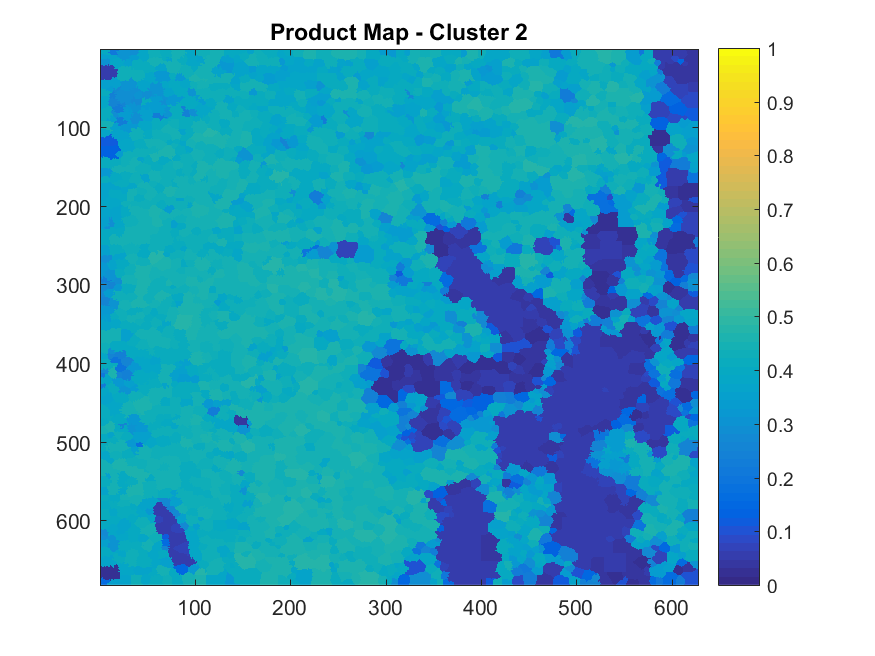}}}
		\subfigure[ PFCM Cluster 3 ]{\resizebox{!}{.95in}{\includegraphics[trim={1.6cm 1cm 3cm .75cm},clip]{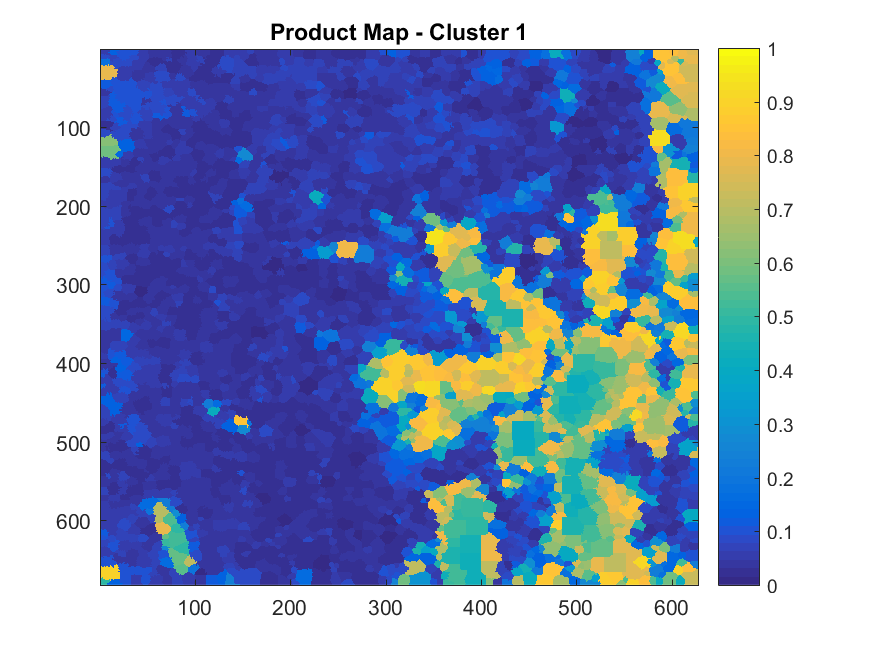}}}\\
     		\caption[]{ Clustering results of Fig. 2 given by the PFLICM (a-c), FLICM (d-f), and PFCM (g-i) algorithms. Clusters have been manually aligned for easy comparison. }
		\label{fig:im4}		}
\end{figure}

\begin{figure}[!t]
	\center {
			\subfigure[ PFLICM Cluster 1 Typicality Map ]{\resizebox{!}{.95in}{\includegraphics[trim={1.6cm 1cm 3cm .75cm},clip]{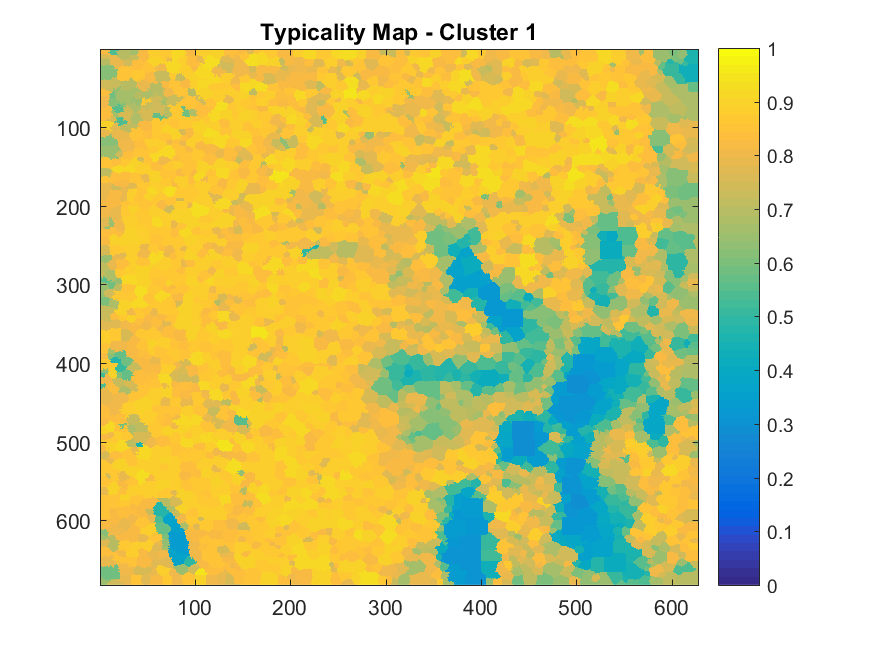}}}
		\subfigure[ PFLICM Cluster 2 Typicality Map ]{\resizebox{!}{.95in}{\includegraphics[trim={1.6cm 1cm 3cm .75cm},clip]{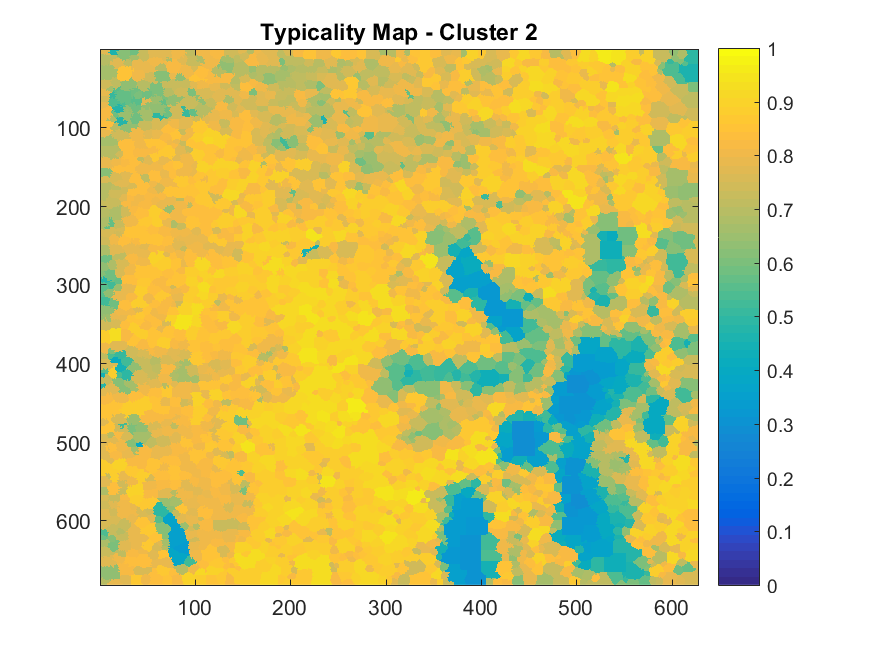}}}
		\subfigure[ PFLICM Cluster 3 Typicality Map ]{\resizebox{!}{.95in}{\includegraphics[trim={1.6cm 1cm 3cm .75cm},clip]{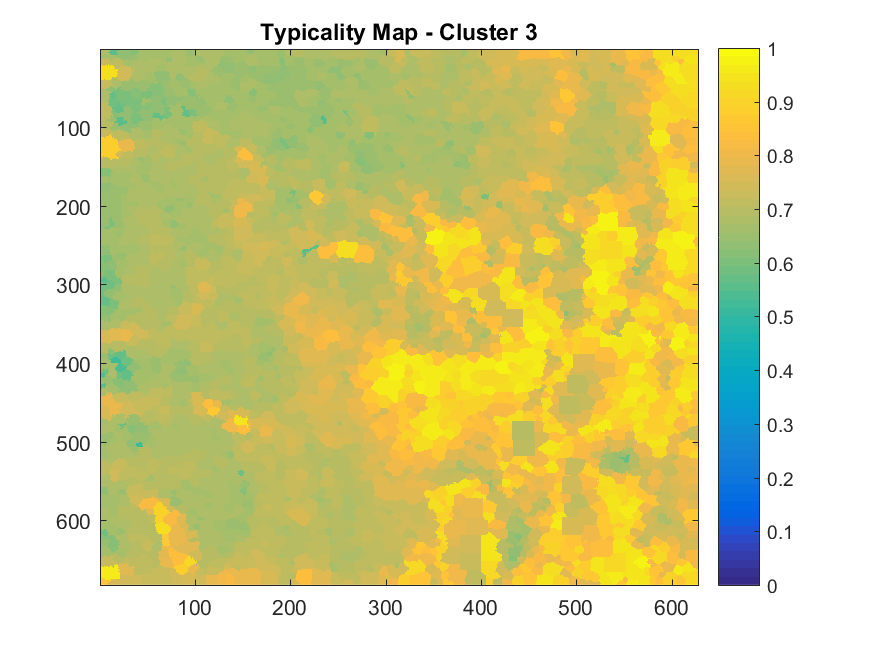}}}\\
		\subfigure[ FLICM Cluster 1 Membership Map ]{\resizebox{!}{.95in}{\includegraphics[trim={1.6cm 1cm 3cm .75cm},clip]{Figures/ims/im4/FLICM_C1.png}}}
		\subfigure[ FLICM Cluster 2 Membership Map ]{\resizebox{!}{.95in}{\includegraphics[trim={1.6cm 1cm 3cm .75cm},clip]{Figures/ims/im4/FLICM_C2.png}}}
		\subfigure[ FLICM Cluster 3 Membership Map ]{\resizebox{!}{.95in}{\includegraphics[trim={1.6cm 1cm 3cm .75cm},clip]{Figures/ims/im4/FLICM_C3.png}}}\\
		\subfigure[ PFCM Cluster 1 Typicality Map ]{\resizebox{!}{.95in}{\includegraphics[trim={1.6cm 1cm 3cm .75cm},clip]{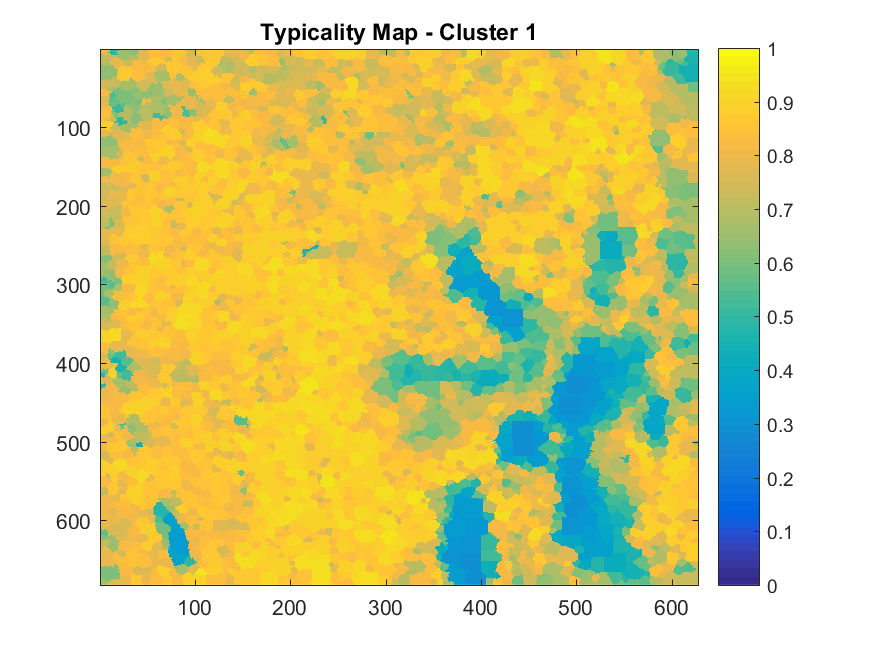}}}
		\subfigure[ PFCM Cluster 2 Typicality Map ]{\resizebox{!}{.95in}{\includegraphics[trim={1.6cm 1cm 3cm .75cm},clip]{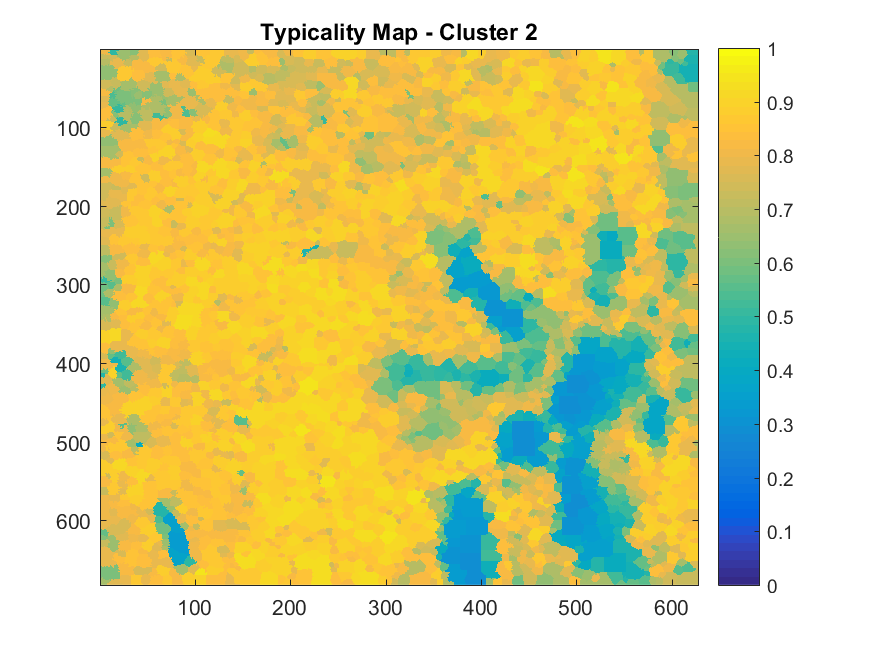}}}
		\subfigure[ PFCM Cluster 3 Typicality Map ]{\resizebox{!}{.95in}{\includegraphics[trim={1.6cm 1cm 3cm .75cm},clip]{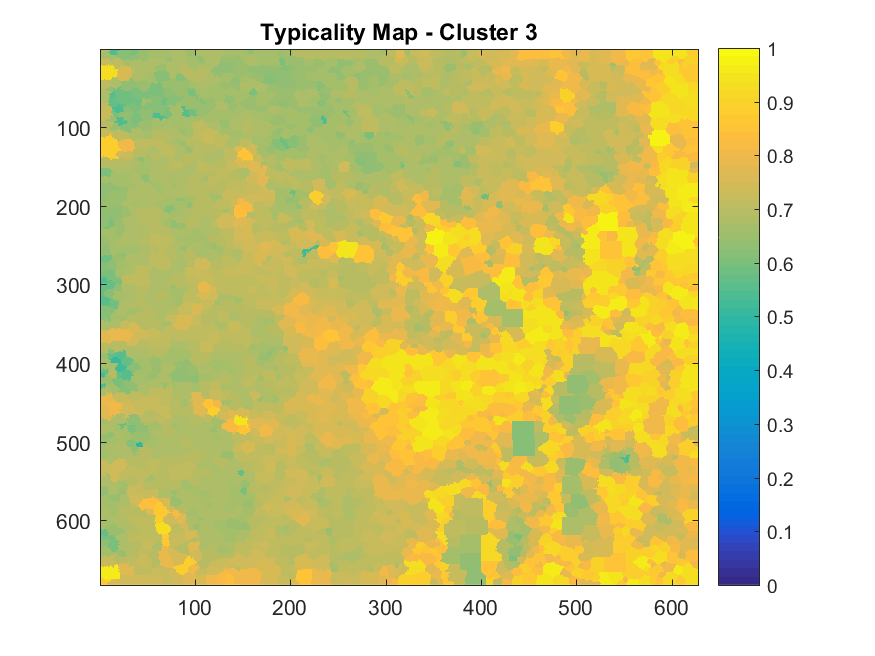}}}
		\caption[]{ Clustering results on image (a) from Figure \ref{fig:im4} displaying the typicalities of PFLICM (a-c), memberships of FLICM (d-f), and typicalities of PFCM (g-i). } 
		\label{fig:im4a}		}
\end{figure}

\begin{figure}[!t]
	\center{
			\subfigure[ Sub-Image containing an outlier object]{\resizebox{!}{2.5in}{\includegraphics[trim={1cm 0cm 1cm .75cm},clip]{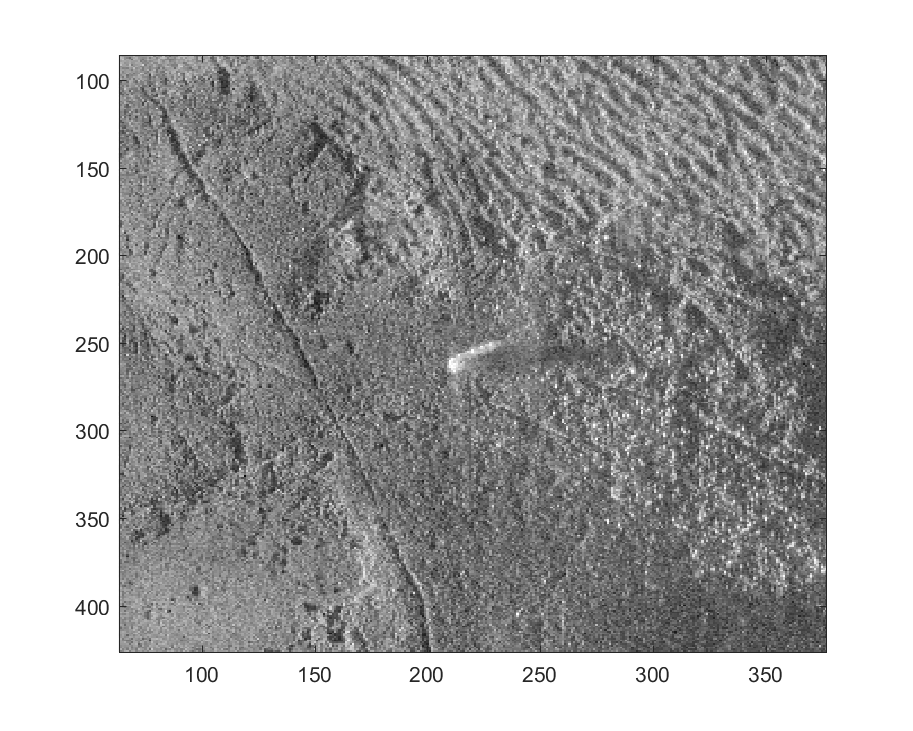}}}\\
		\subfigure[  Outlier - Typicality Cluster 1 ]{\resizebox{!}{.95in}{\includegraphics[trim={1.6cm 1cm 3cm .75cm},clip]{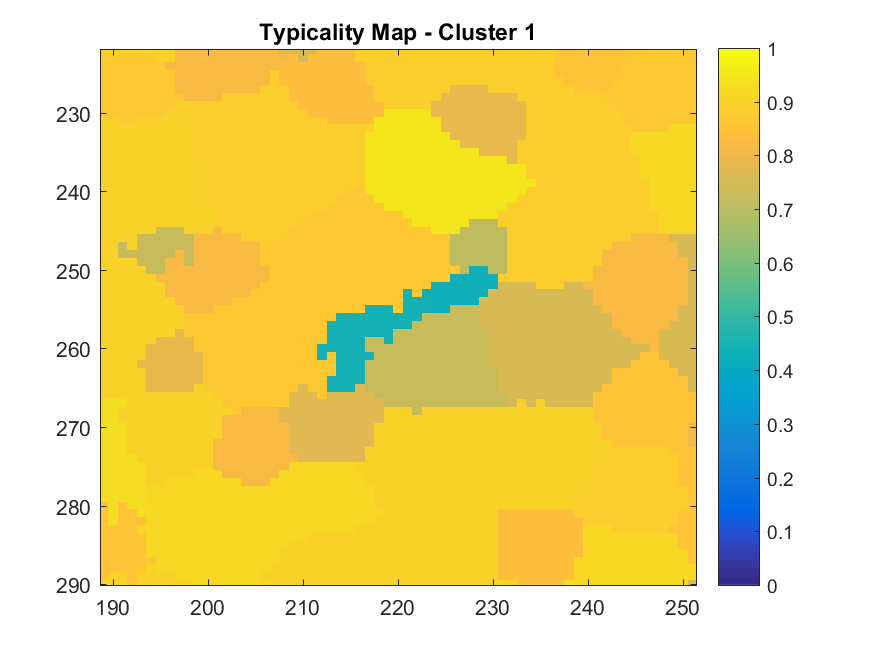}}}
		\subfigure[ Outlier - Typicality Cluster 2 ]{\resizebox{!}{.95in}{\includegraphics[trim={1.6cm 1cm 3cm .75cm},clip]{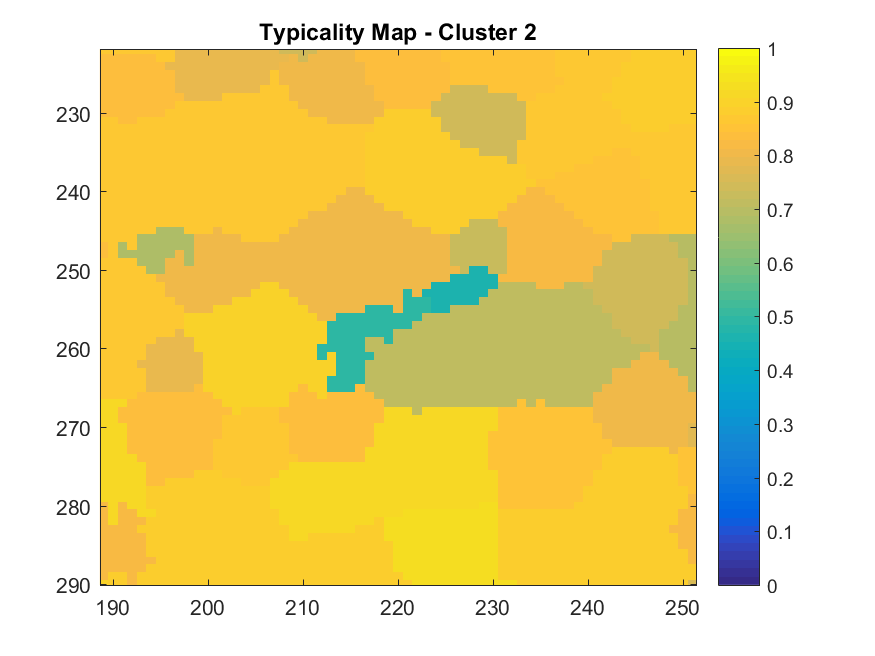}}}
		\subfigure[  Outlier - Typicality Cluster 3 ]{\resizebox{!}{.95in}{\includegraphics[trim={1.6cm 1cm 3cm .75cm},clip]{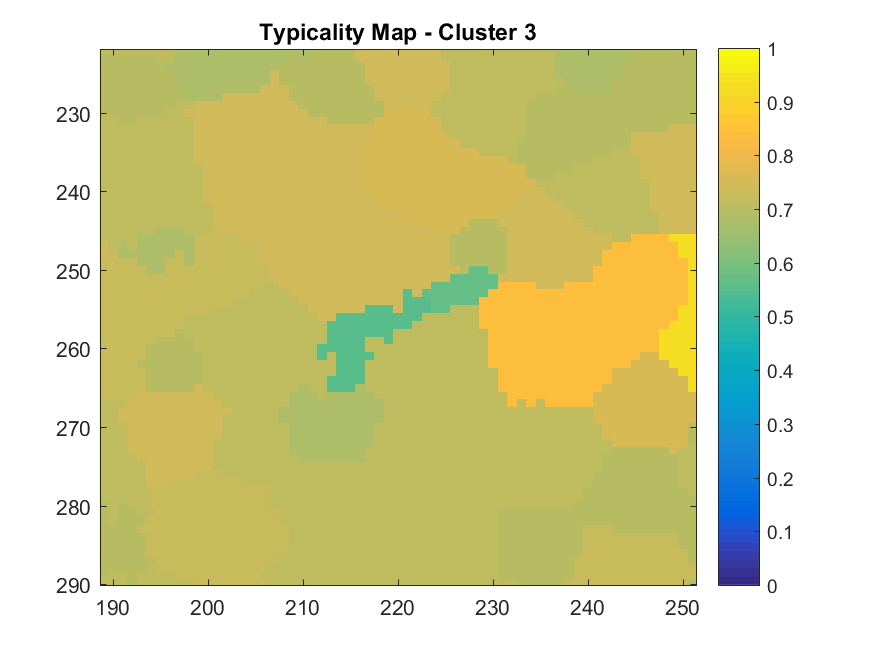}}}\\
		\subfigure[  Outlier - Membership Cluster 1 ]{\resizebox{!}{.95in}{\includegraphics[trim={1.6cm 1cm 3cm .75cm},clip]{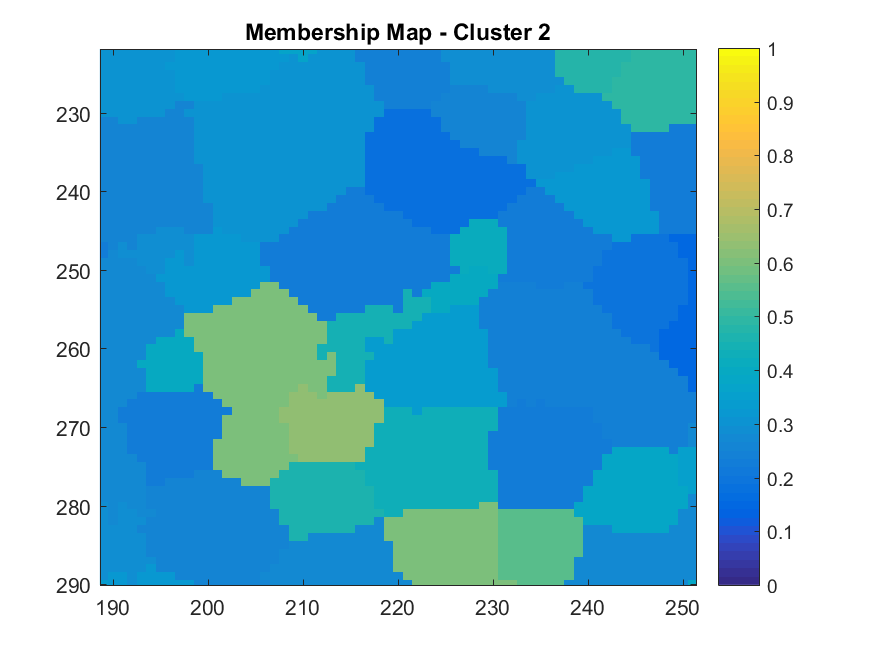}}}
		\subfigure[ Outlier - Membership Cluster 2 ]{\resizebox{!}{.95in}{\includegraphics[trim={1.6cm 1cm 3cm .75cm},clip]{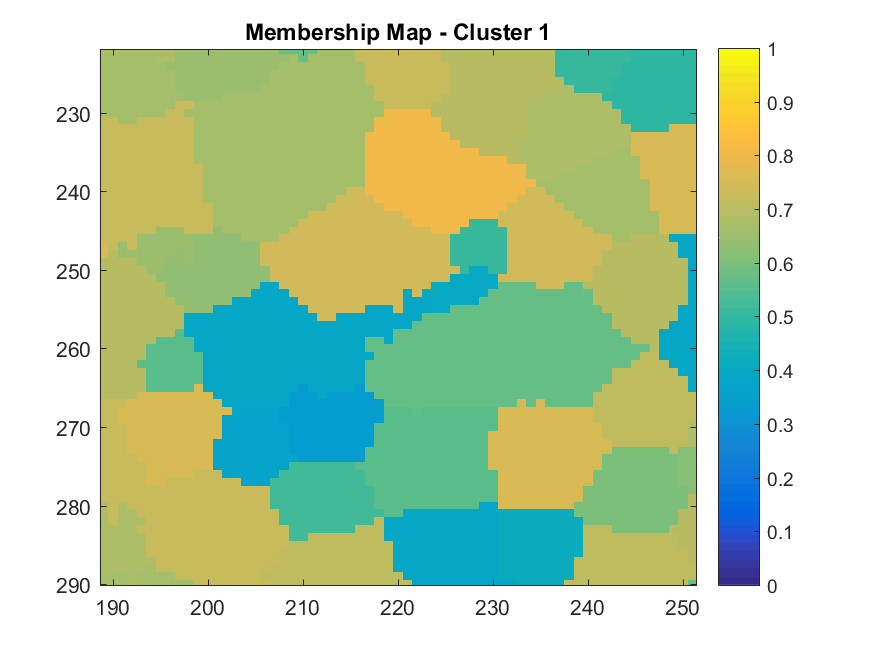}}}
		\subfigure[ Outlier - Membership Cluster 3 ]{\resizebox{!}{.95in}{\includegraphics[trim={1.6cm 1cm 3cm .75cm},clip]{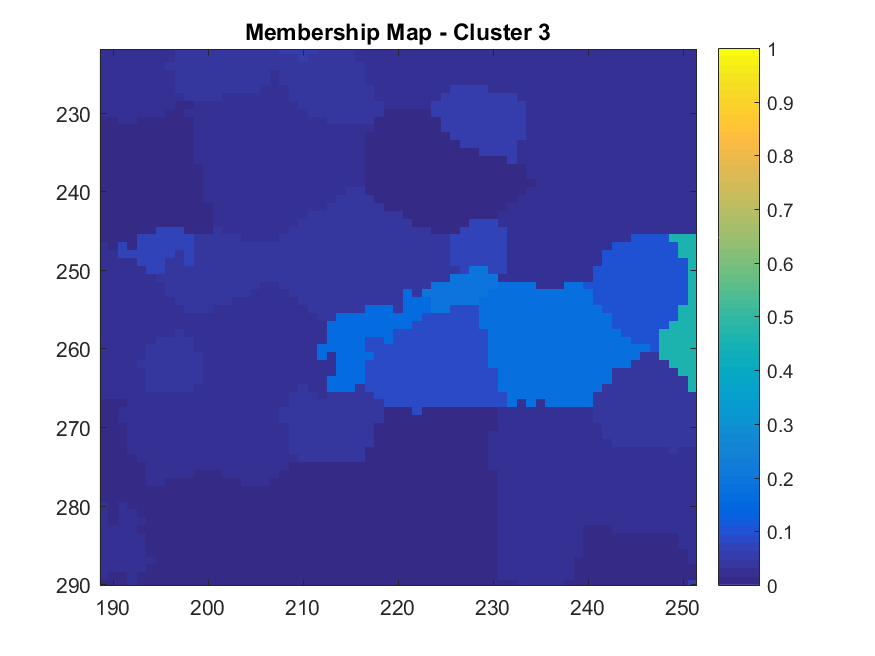}}}\\
     		\caption[]{(a) Sub-Image containing an outlier obtained from the image (in Fig. 2. (b-d) show a zoomed in area around the outlier of the typicality maps from Figure \ref{fig:im4a} centered on the outlier. (e-g) show a zoomed in area around the outlier on the membership maps from Figure \ref{fig:im4a} centered on the outlier. }
		\label{fig:im4b}		}
\end{figure}

\begin{figure}[!h]
	\center{
		\subfigure[ Original Image ]{\resizebox{!}{2.5in}{\includegraphics[trim={1cm 0cm 1cm .75cm},clip]{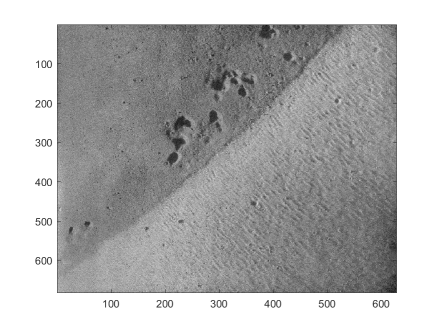}}}\\
		\subfigure[ PFLICM Cluster 1 ]{\resizebox{!}{.95in}{\includegraphics[trim={1.6cm 1cm 3cm .75cm},clip]{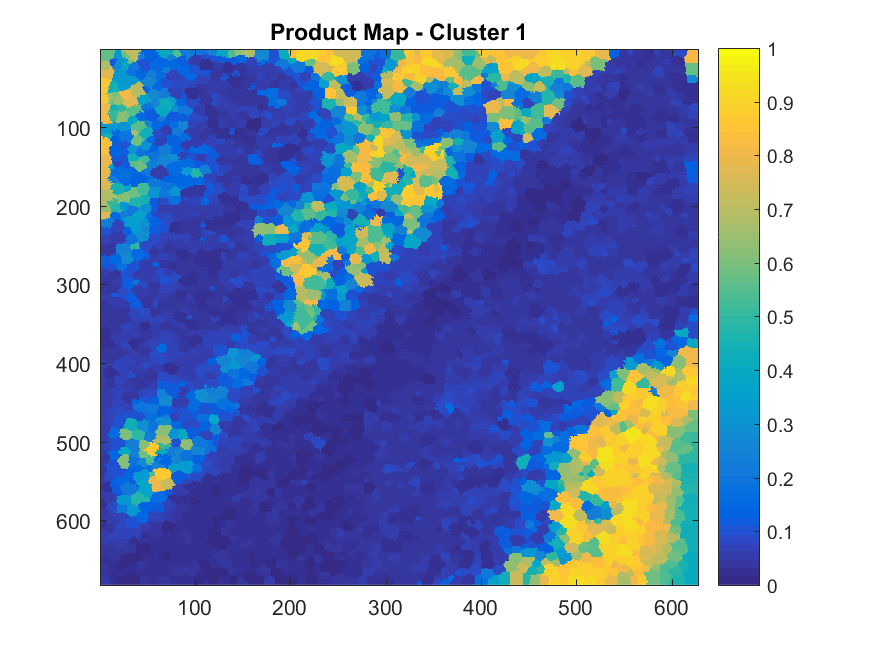}}}
		\subfigure[ PFLICM Cluster 2 ]{\resizebox{!}{.95in}{\includegraphics[trim={1.6cm 1cm 3cm .75cm},clip]{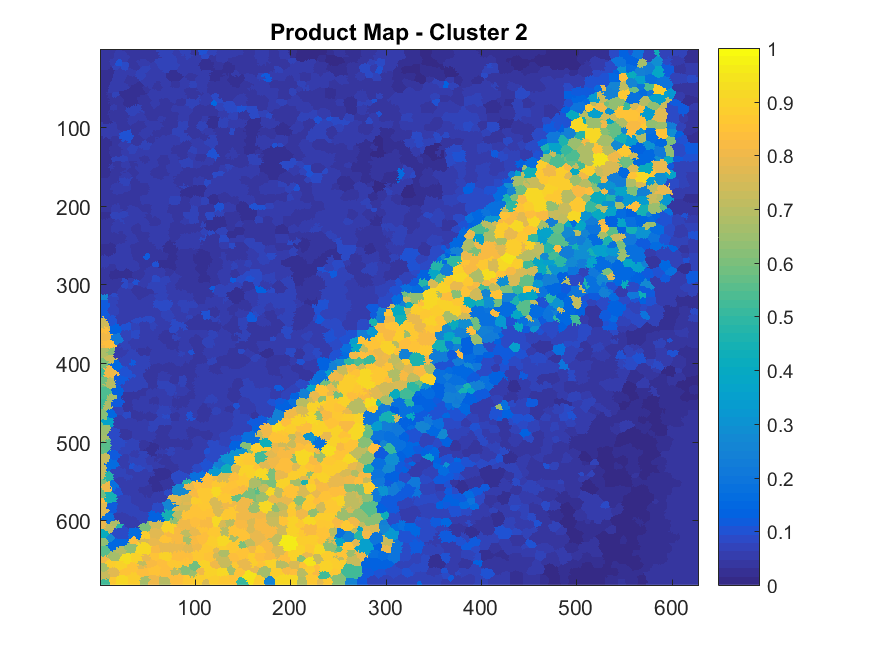}}}
		\subfigure[ PFLICM Cluster 3 ]{\resizebox{!}{.95in}{\includegraphics[trim={1.6cm 1cm 3cm .75cm},clip]{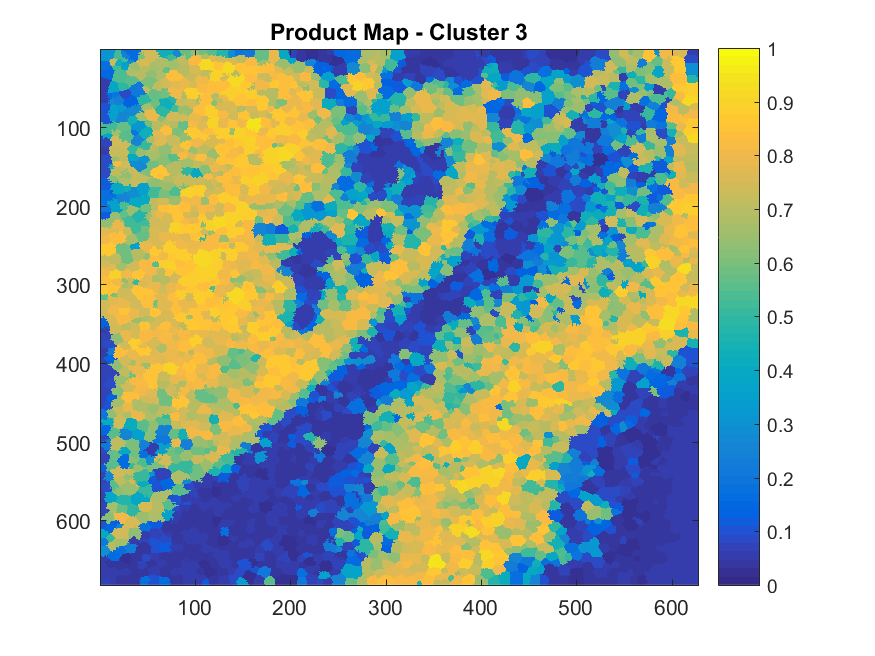}}}\\
		\subfigure[ FLICM Cluster 1 ]{\resizebox{!}{.95in}{\includegraphics[trim={1.6cm 1cm 3cm .75cm},clip]{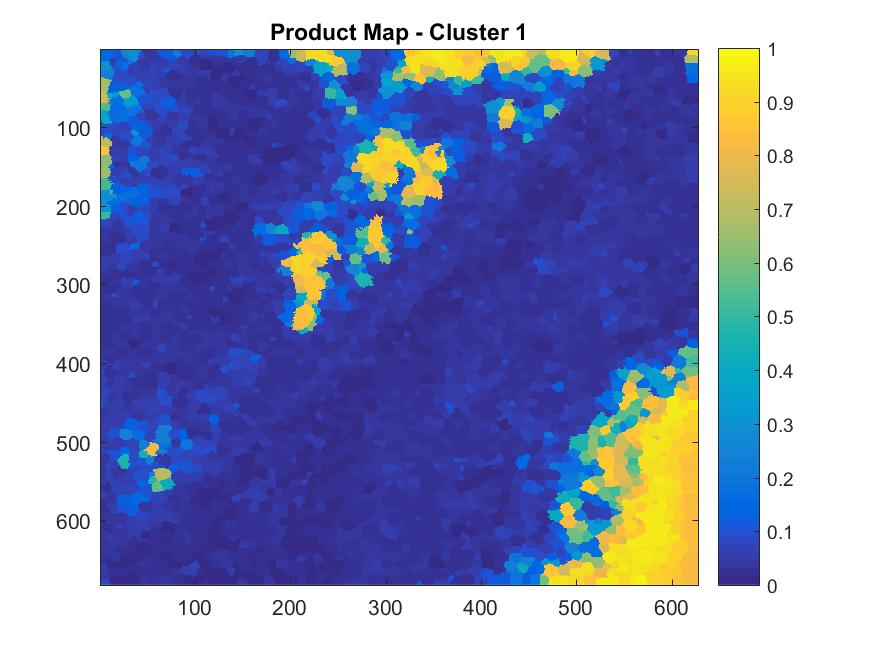}}}
		\subfigure[ FLICM Cluster 2 ]{\resizebox{!}{.95in}{\includegraphics[trim={1.6cm 1cm 3cm .75cm},clip]{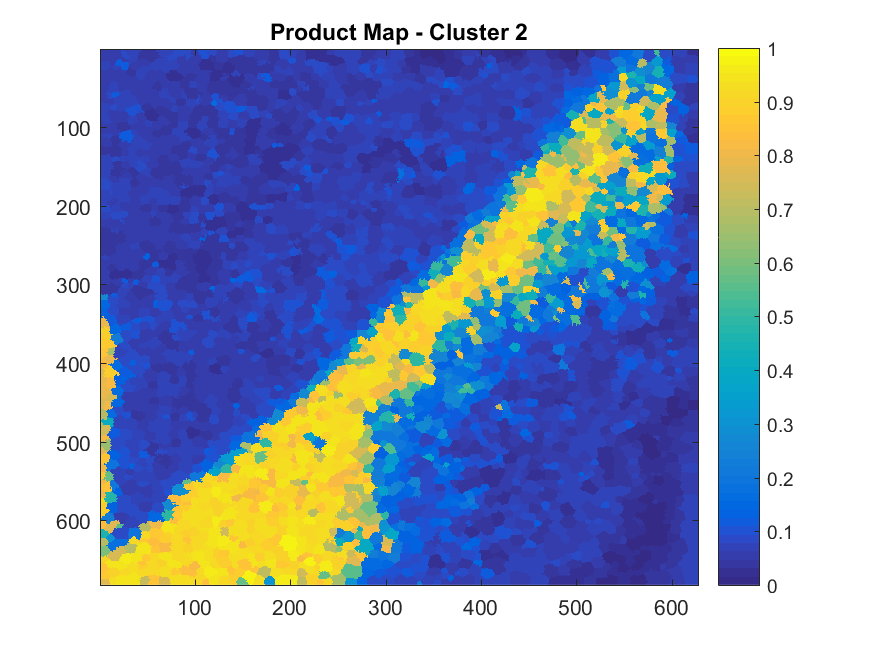}}}
		\subfigure[ FLICM Cluster 3 ]{\resizebox{!}{.95in}{\includegraphics[trim={1.6cm 1cm 3cm .75cm},clip]{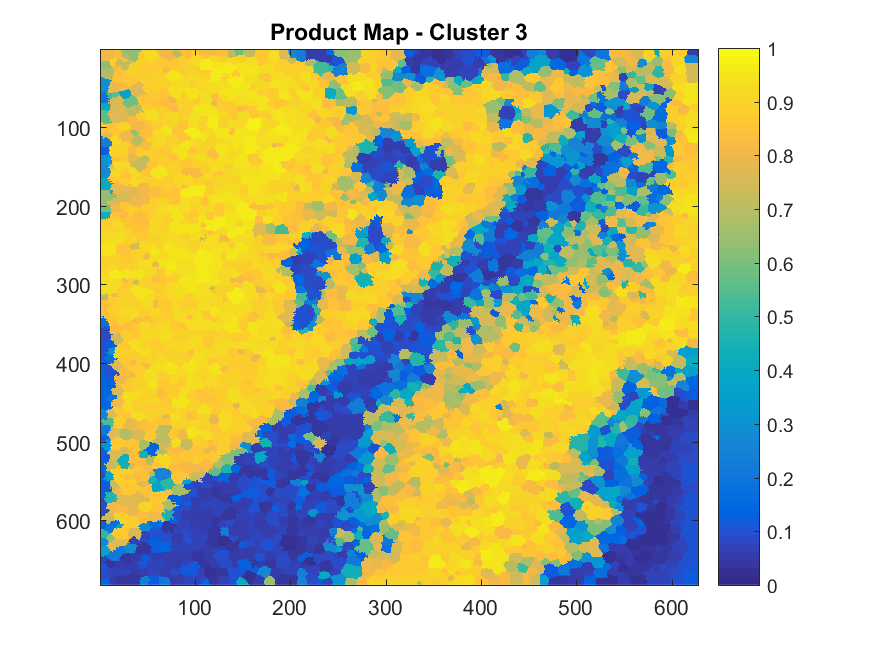}}}\\
		\subfigure[ PFCM Cluster 1 ]{\resizebox{!}{.95in}{\includegraphics[trim={1.6cm 1cm 3cm .75cm},clip]{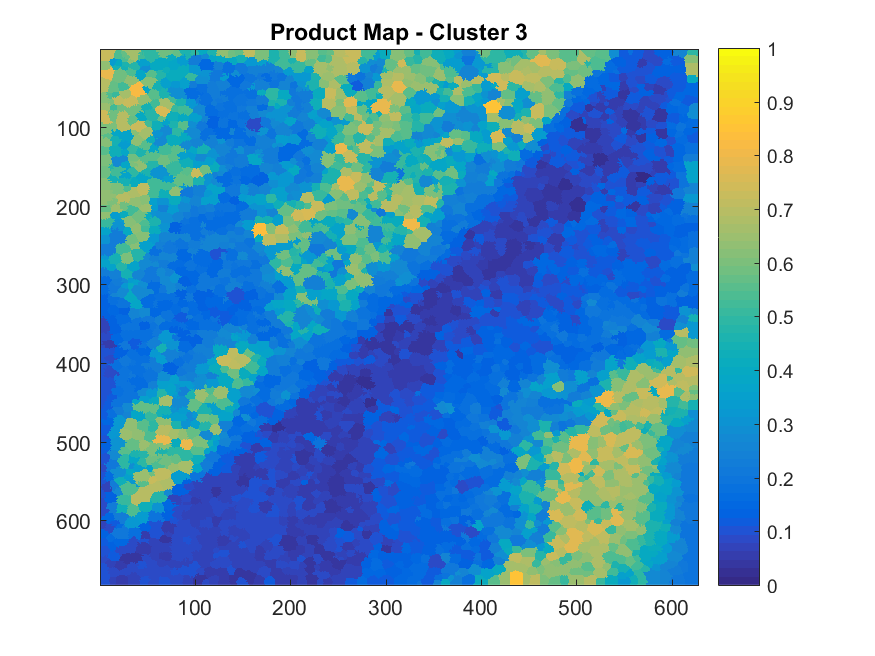}}}
		\subfigure[ PFCM Cluster 2 ]{\resizebox{!}{.95in}{\includegraphics[trim={1.6cm 1cm 3cm .75cm},clip]{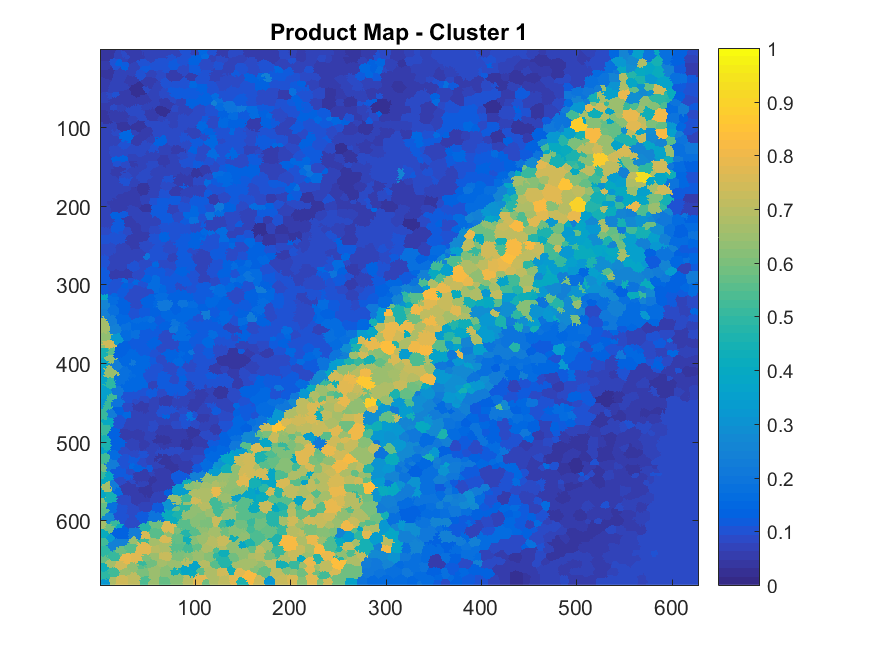}}}
		\subfigure[ PFCM Cluster 3 ]{\resizebox{!}{.95in}{\includegraphics[trim={1.6cm 1cm 3cm .75cm},clip]{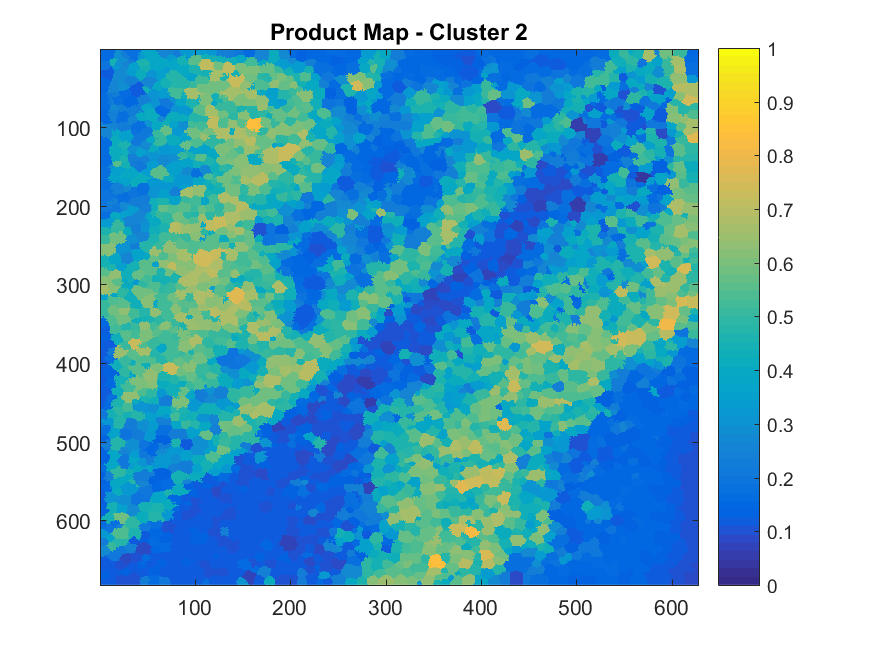}}}\\
     		\caption[]{ (a) Image containing sand ripples, hard-packed sand, and holes. Clustering results of image (a) given by the PFLICM (b-d), FLICM (e-g), and PFCM (h-j) algorithms. Clusters have been manually aligned for easy comparison. }
		\label{fig:im1}		}
\end{figure}

\begin{figure}[!t]
	\center{
		\subfigure[ Original Image ]{\resizebox{!}{2.5in}{\includegraphics[trim={1cm 0cm 1cm .75cm},clip]{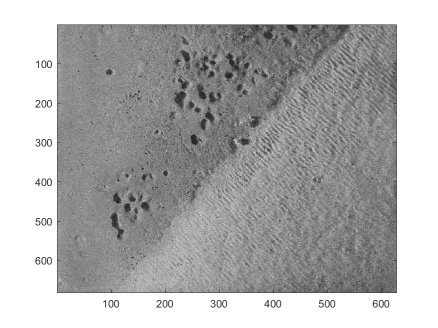}}}\\
		\subfigure[ PFLICM Cluster 1 ]{\resizebox{!}{.95in}{\includegraphics[trim={1.6cm 1cm 3cm .75cm},clip]{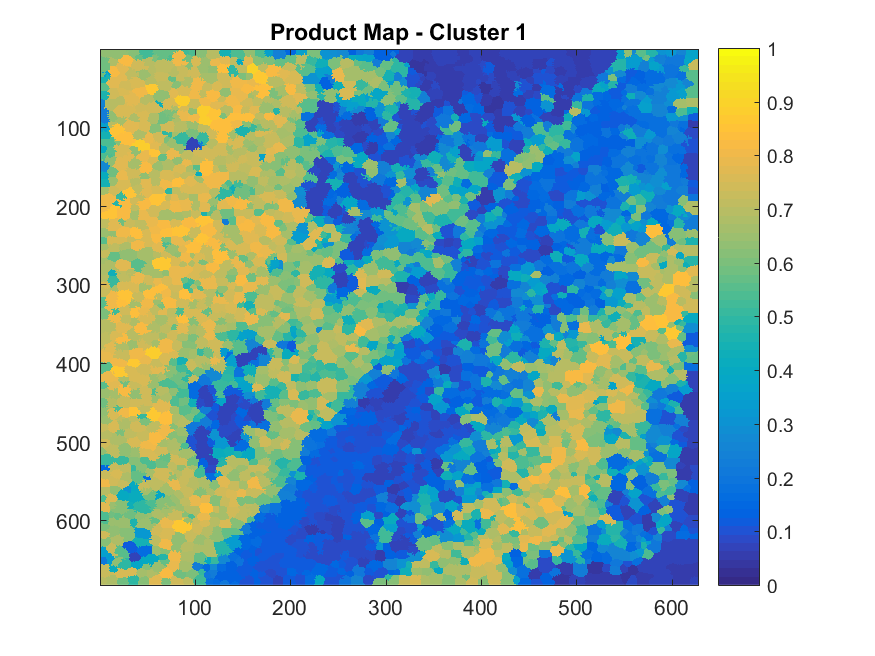}}}
		\subfigure[ PFLICM Cluster 2 ]{\resizebox{!}{.95in}{\includegraphics[trim={1.6cm 1cm 3cm .75cm},clip]{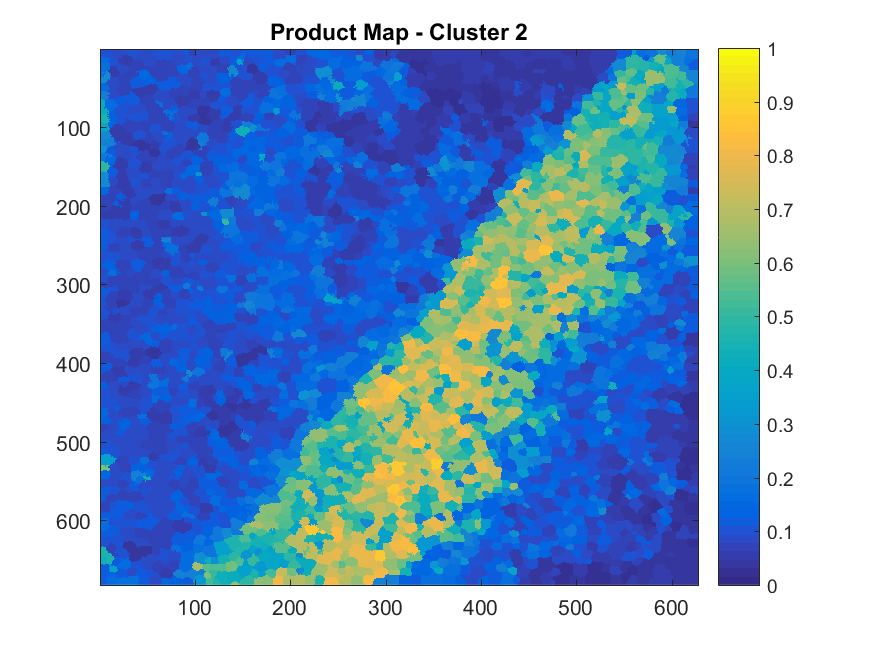}}}
		\subfigure[ PFLICM Cluster 3 ]{\resizebox{!}{.95in}{\includegraphics[trim={1.6cm 1cm 3cm .75cm},clip]{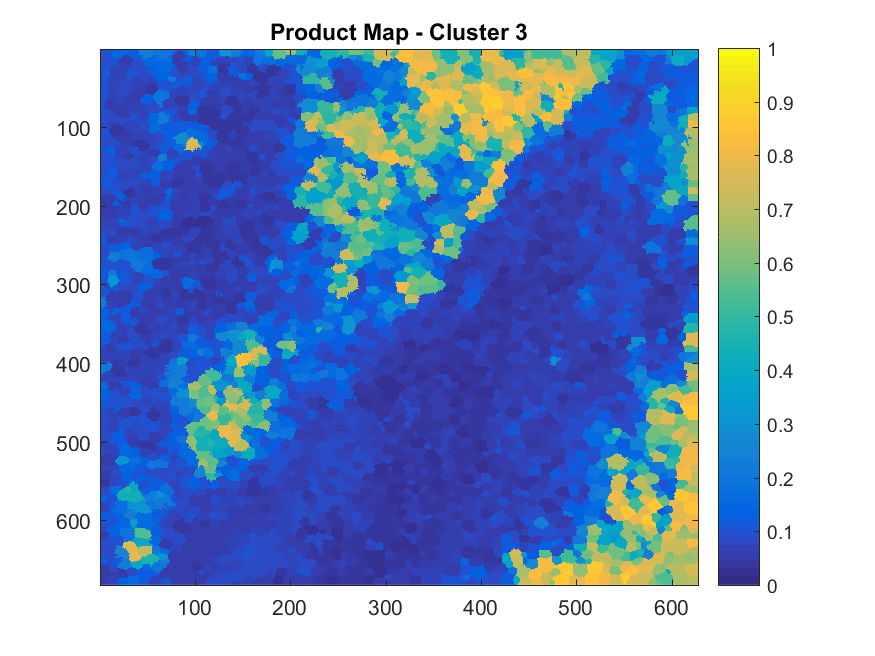}}}\\
		\subfigure[ FLICM Cluster 1 ]{\resizebox{!}{.95in}{\includegraphics[trim={1.6cm 1cm 3cm .75cm},clip]{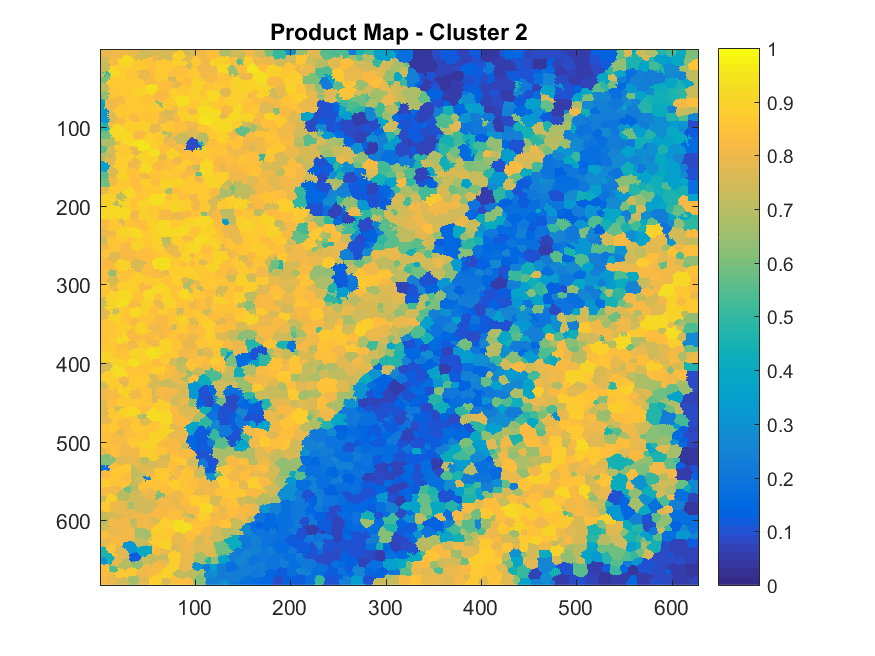}}}
		\subfigure[ FLICM Cluster 2 ]{\resizebox{!}{.95in}{\includegraphics[trim={1.6cm 1cm 3cm .75cm},clip]{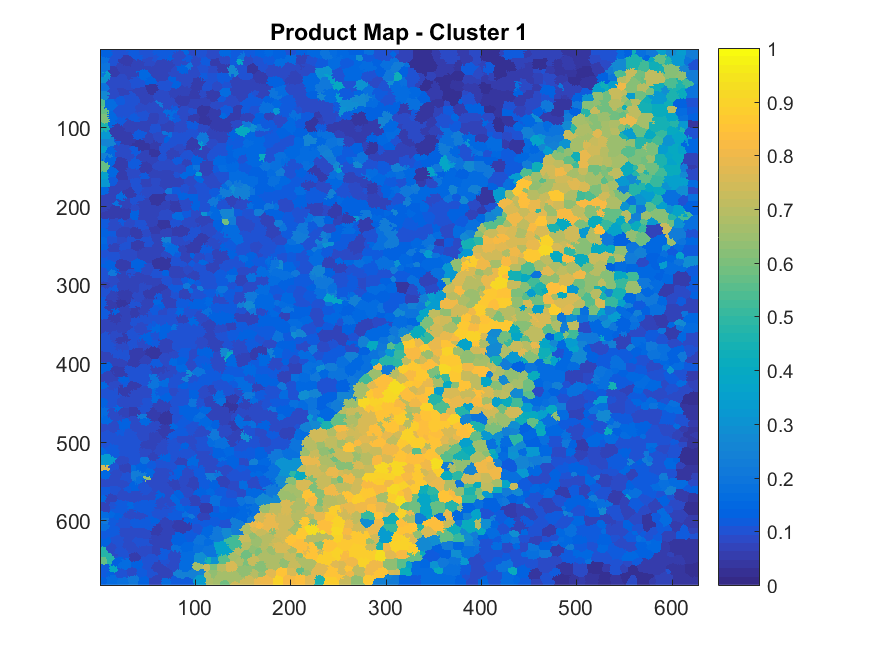}}}
		\subfigure[ FLICM Cluster 3 ]{\resizebox{!}{.95in}{\includegraphics[trim={1.6cm 1cm 3cm .75cm},clip]{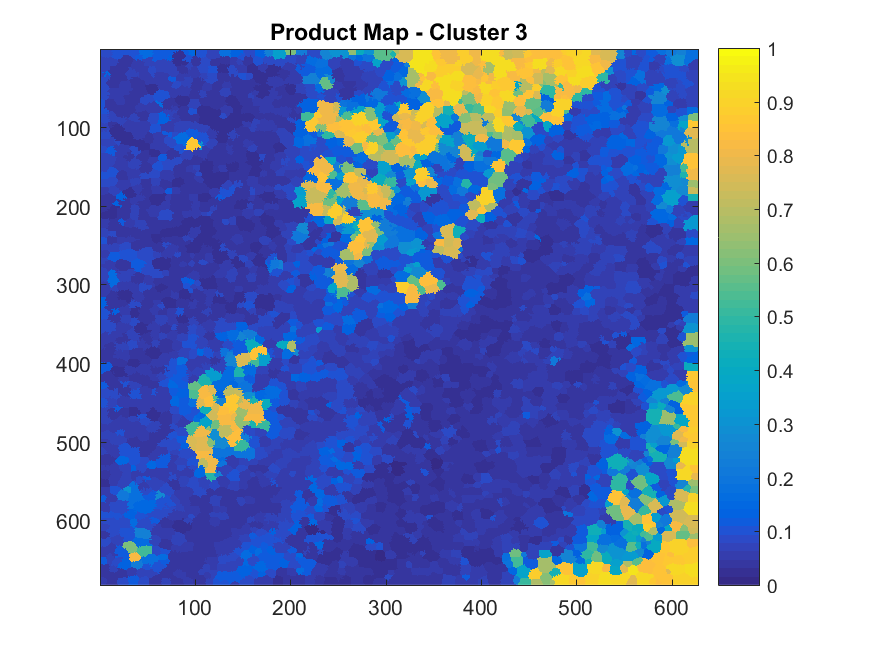}}}\\
		\subfigure[ PFCM Cluster 1 ]{\resizebox{!}{.95in}{\includegraphics[trim={1.6cm 1cm 3cm .75cm},clip]{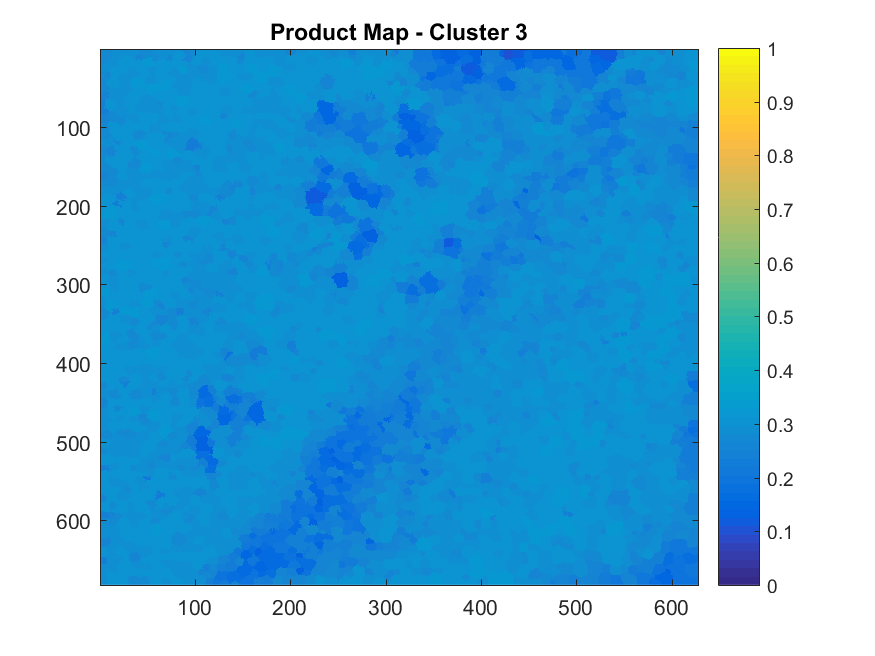}}}
		\subfigure[ PFCM Cluster 2 ]{\resizebox{!}{.95in}{\includegraphics[trim={1.6cm 1cm 3cm .75cm},clip]{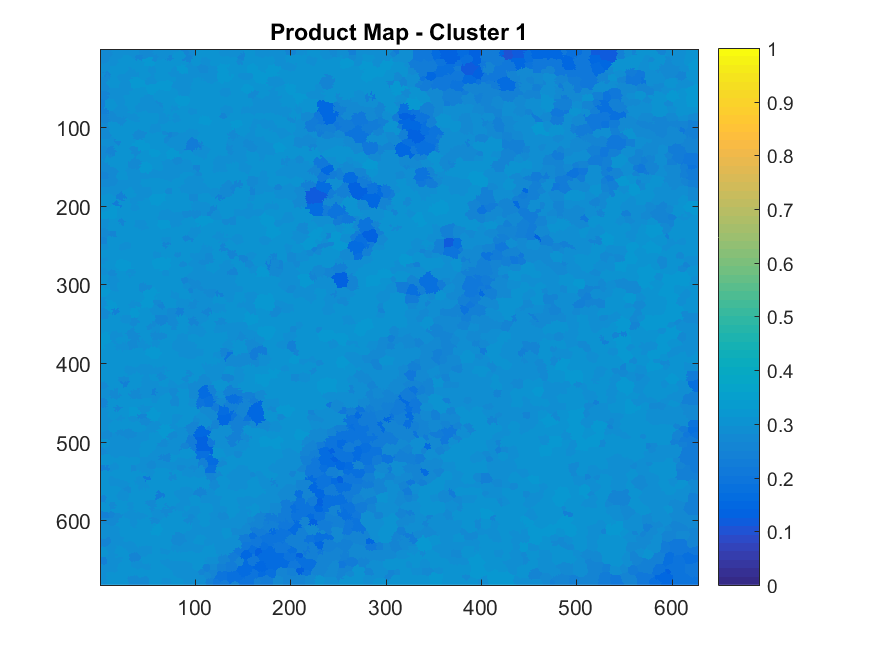}}}
		\subfigure[ PFCM Cluster 3 ]{\resizebox{!}{.95in}{\includegraphics[trim={1.6cm 1cm 3cm .75cm},clip]{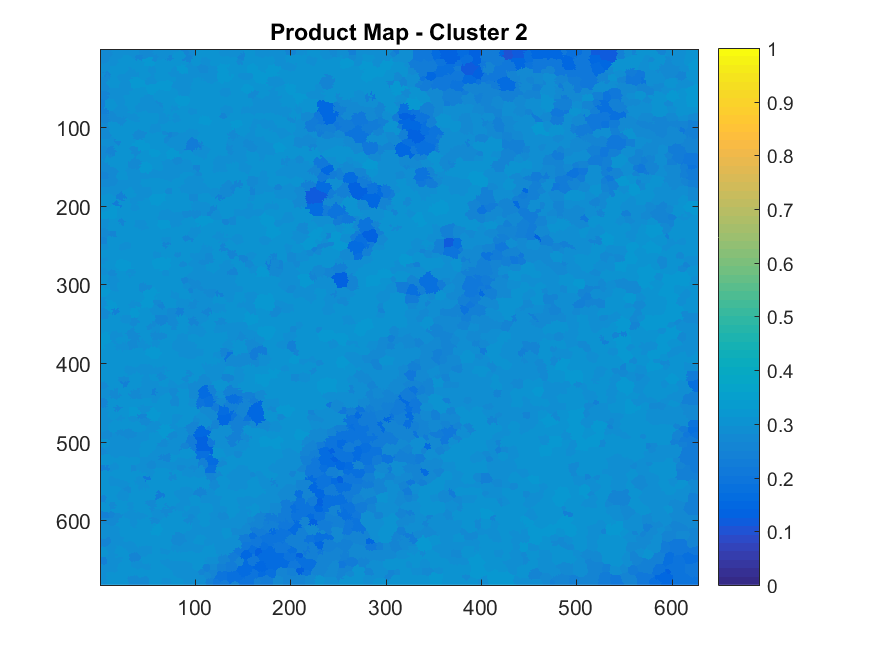}}}\\
     		\caption[]{ (a) Image containing sand ripples, hard-packed sand, and holes. Clustering results of image (a) given by the PFLICM (b-d), FLICM (e-g), and PFCM (h-j) algorithms. Clusters have been manually aligned for easy comparison. }
		\label{fig:im2}		}
\end{figure}

\begin{figure}[!t]
	\center{
			\subfigure[ Original Image ]{\resizebox{!}{2.5in}{\includegraphics[trim={1cm 0cm 1cm .75cm},clip]{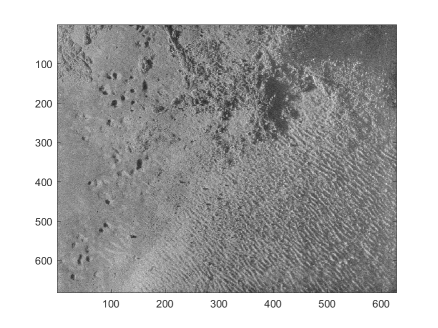}}}\\
		\subfigure[ PFLICM Cluster 1 ]{\resizebox{!}{.95in}{\includegraphics[trim={1.6cm 1cm 3cm .75cm},clip]{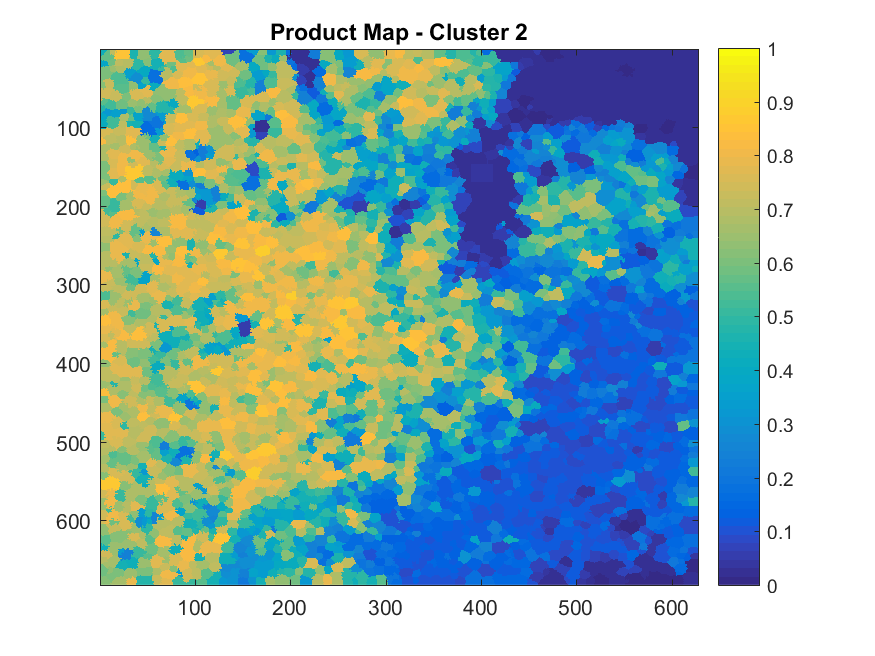}}}
		\subfigure[ PFLICM Cluster 2 ]{\resizebox{!}{.95in}{\includegraphics[trim={1.6cm 1cm 3cm .75cm},clip]{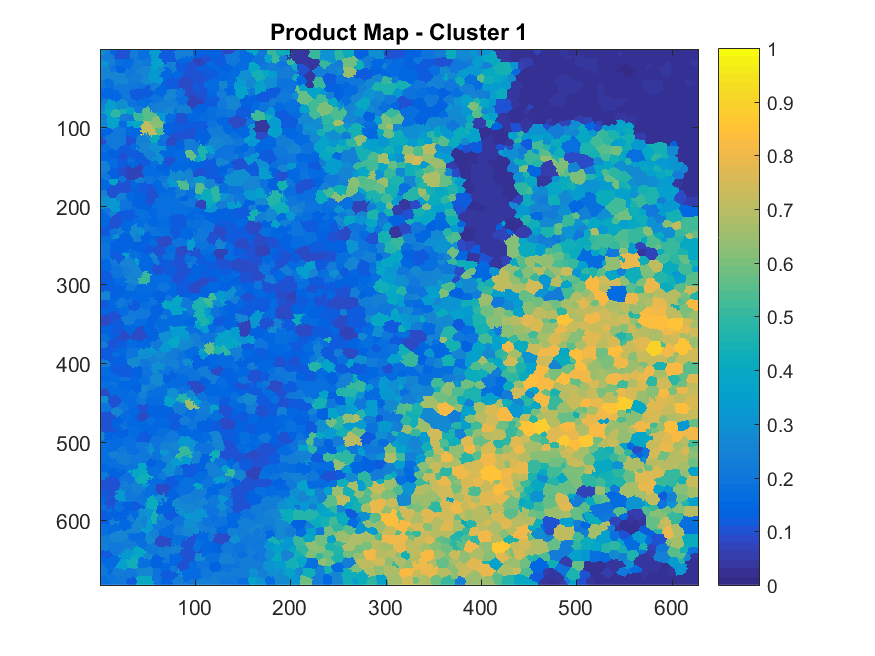}}}
		\subfigure[ PFLICM Cluster 3 ]{\resizebox{!}{.95in}{\includegraphics[trim={1.6cm 1cm 3cm .75cm},clip]{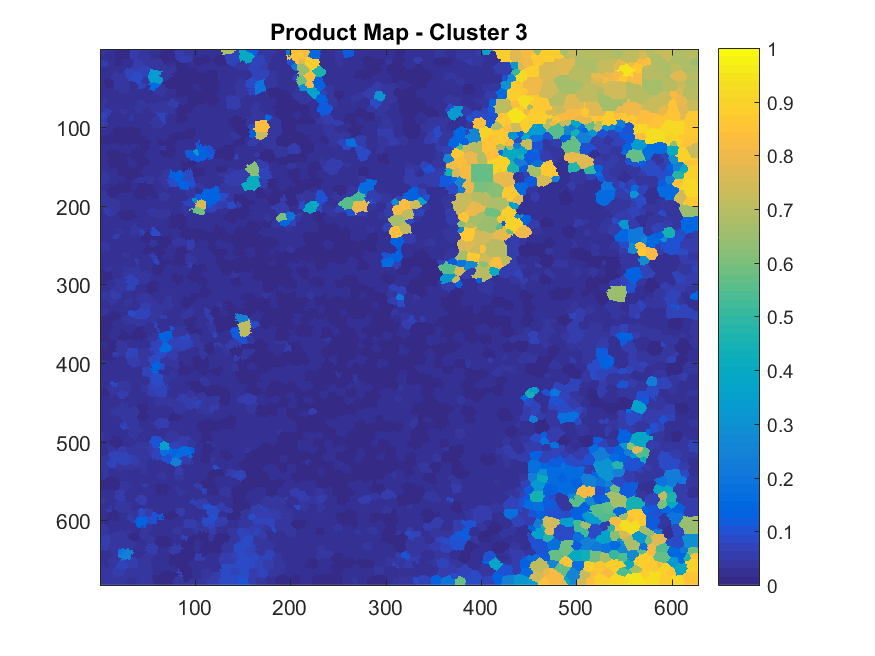}}}\\
		\subfigure[ FLICM Cluster 1 ]{\resizebox{!}{.95in}{\includegraphics[trim={1.6cm 1cm 3cm .75cm},clip]{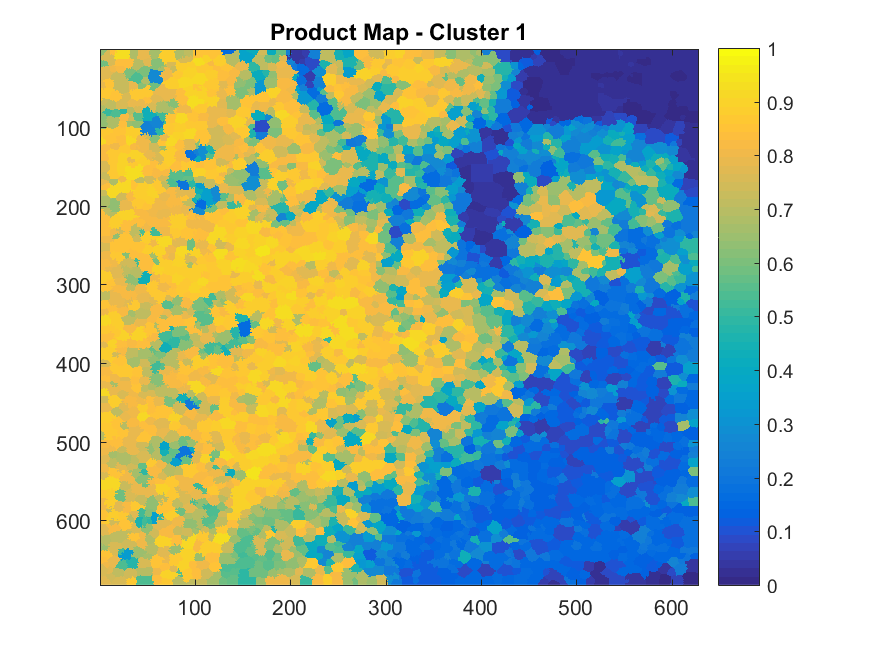}}}
		\subfigure[ FLICM Cluster 2 ]{\resizebox{!}{.95in}{\includegraphics[trim={1.6cm 1cm 3cm .75cm},clip]{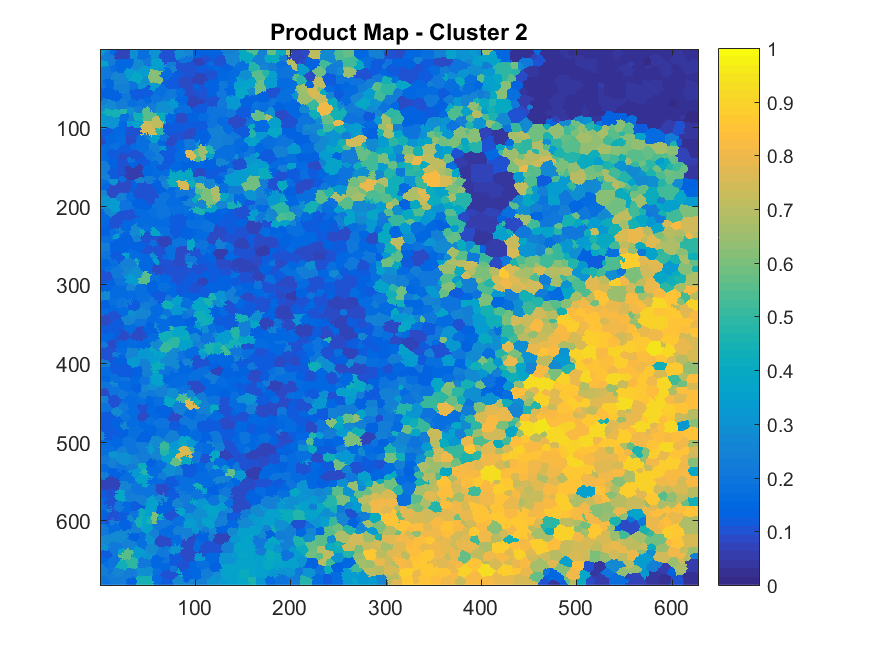}}}
		\subfigure[ FLICM Cluster 3 ]{\resizebox{!}{.95in}{\includegraphics[trim={1.6cm 1cm 3cm .75cm},clip]{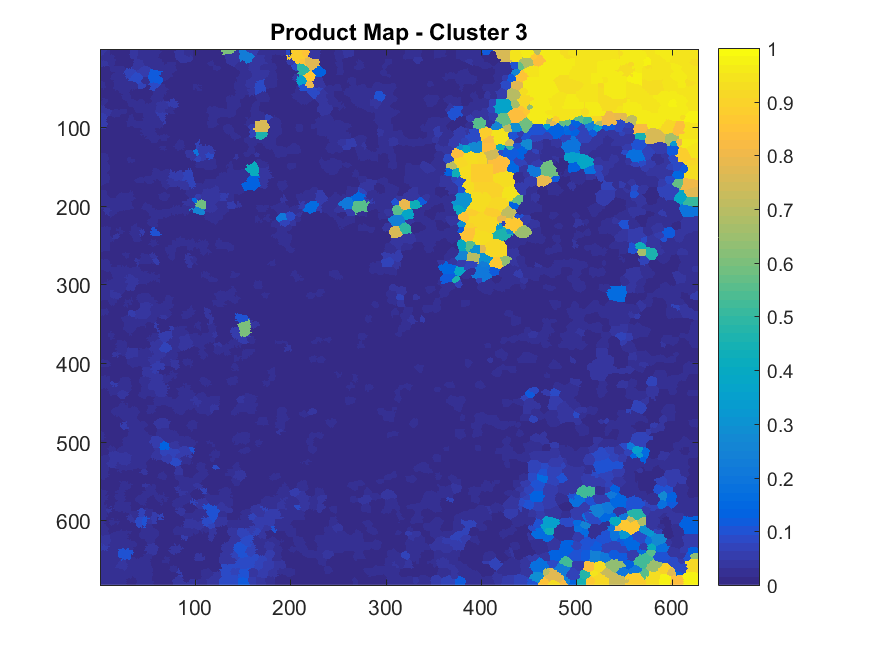}}}\\
		\subfigure[ PFCM Cluster 1 ]{\resizebox{!}{.95in}{\includegraphics[trim={1.6cm 1cm 3cm .75cm},clip]{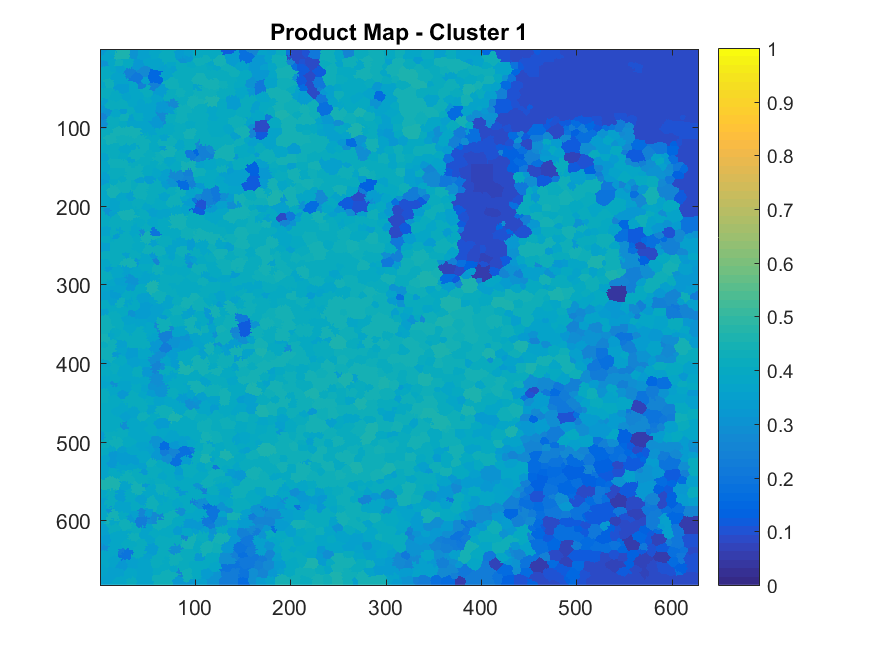}}}
		\subfigure[ PFCM Cluster 2 ]{\resizebox{!}{.95in}{\includegraphics[trim={1.6cm 1cm 3cm .75cm},clip]{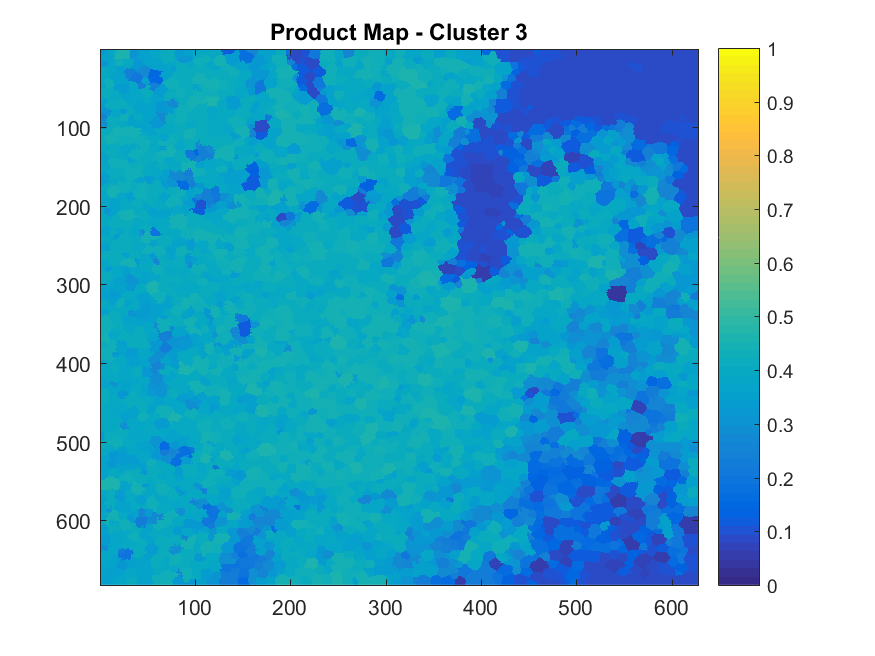}}}
		\subfigure[ PFCM Cluster 3 ]{\resizebox{!}{.95in}{\includegraphics[trim={1.6cm 1cm 3cm .75cm},clip]{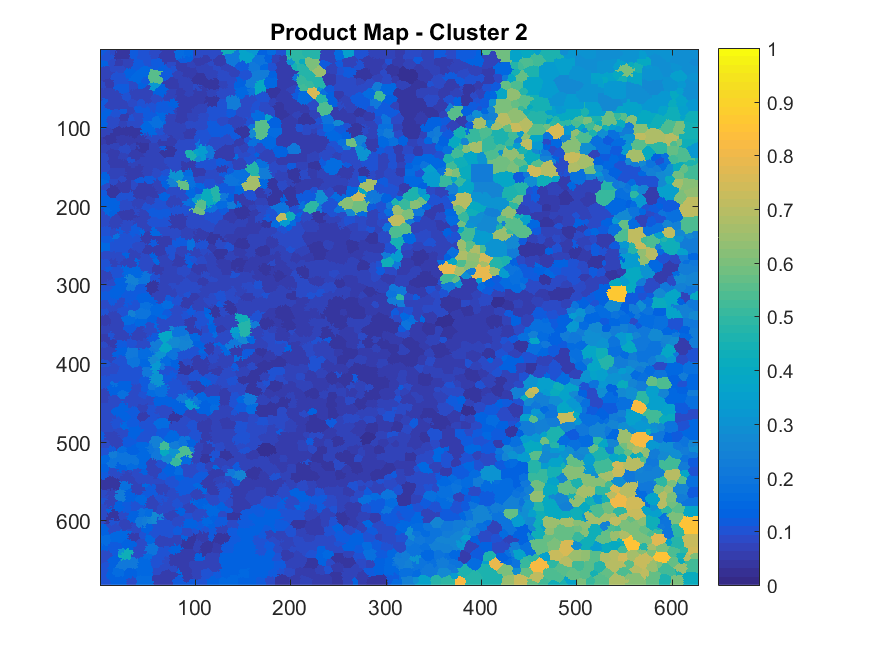}}}\\
     		\caption[]{ (a) Image containing sand ripples, smooth sand, and hilly regions with shadows. Clustering results of image (a) given by the PFLICM (b-d), FLICM (e-g), and PFCM (h-j) algorithms. Clusters have been manually aligned for easy comparison. }
		\label{fig:im3}		}
\end{figure}

\begin{figure}[!t]
	\center{
			\subfigure[ Original Image ]{\resizebox{!}{2.5in}{\includegraphics[trim={1cm 0cm 1cm .75cm},clip]{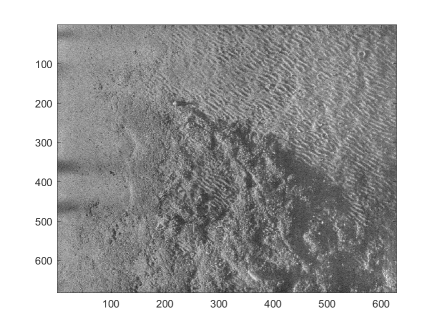}}}\\
		\subfigure[ PFLICM Cluster 1 ]{\resizebox{!}{.95in}{\includegraphics[trim={1.6cm 1cm 3cm .75cm},clip]{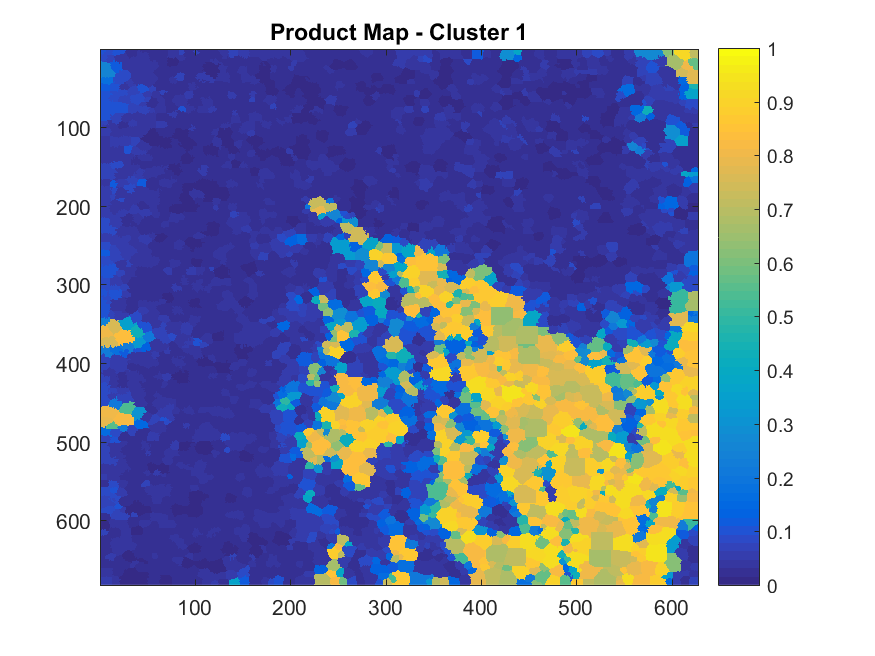}}}
		\subfigure[ PFLICM Cluster 2 ]{\resizebox{!}{.95in}{\includegraphics[trim={1.6cm 1cm 3cm .75cm},clip]{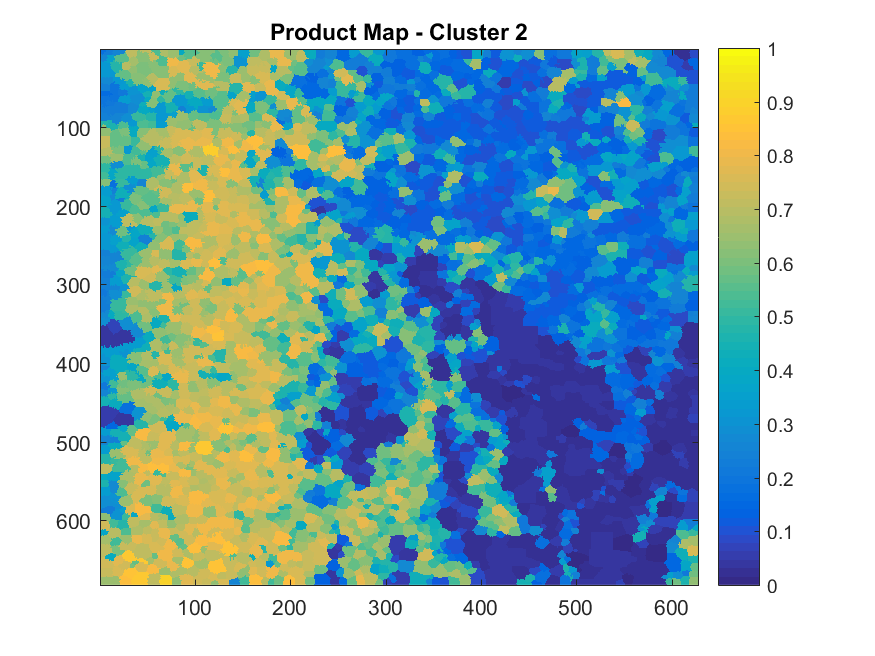}}}
		\subfigure[ PFLICM Cluster 3 ]{\resizebox{!}{.95in}{\includegraphics[trim={1.6cm 1cm 3cm .75cm},clip]{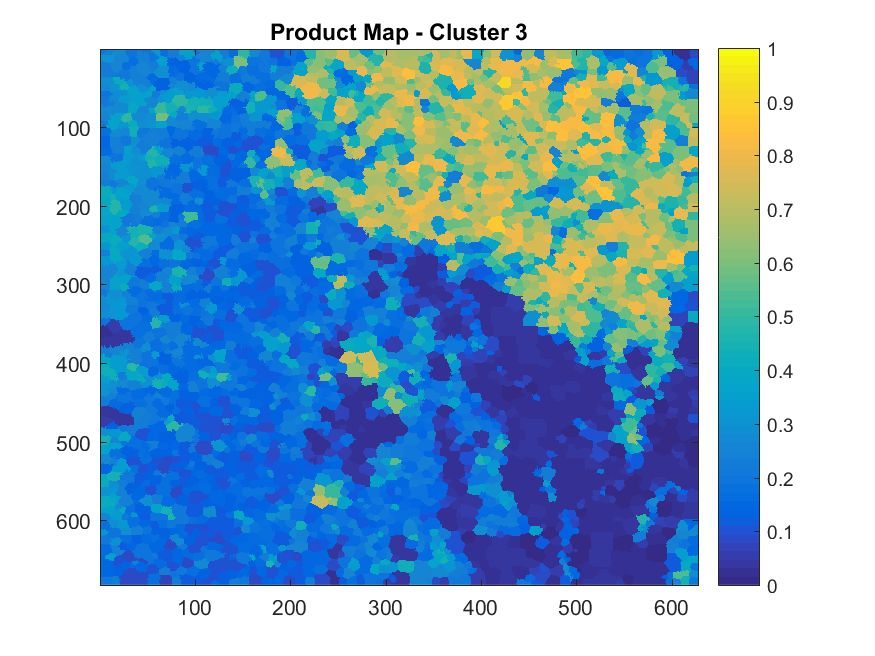}}}\\
		\subfigure[ FLICM Cluster 1 ]{\resizebox{!}{.95in}{\includegraphics[trim={1.6cm 1cm 3cm .75cm},clip]{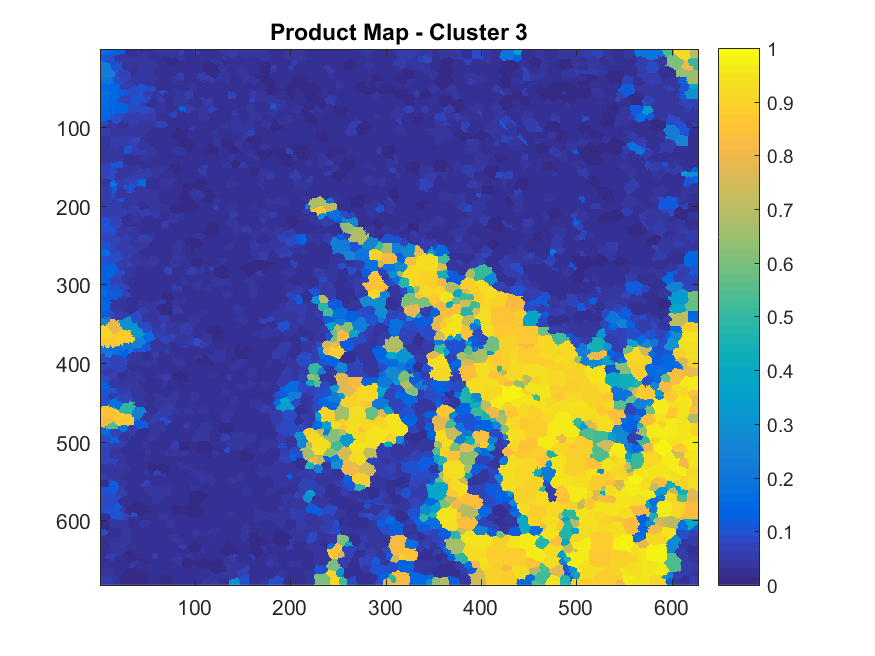}}}
		\subfigure[ FLICM Cluster 2 ]{\resizebox{!}{.95in}{\includegraphics[trim={1.6cm 1cm 3cm .75cm},clip]{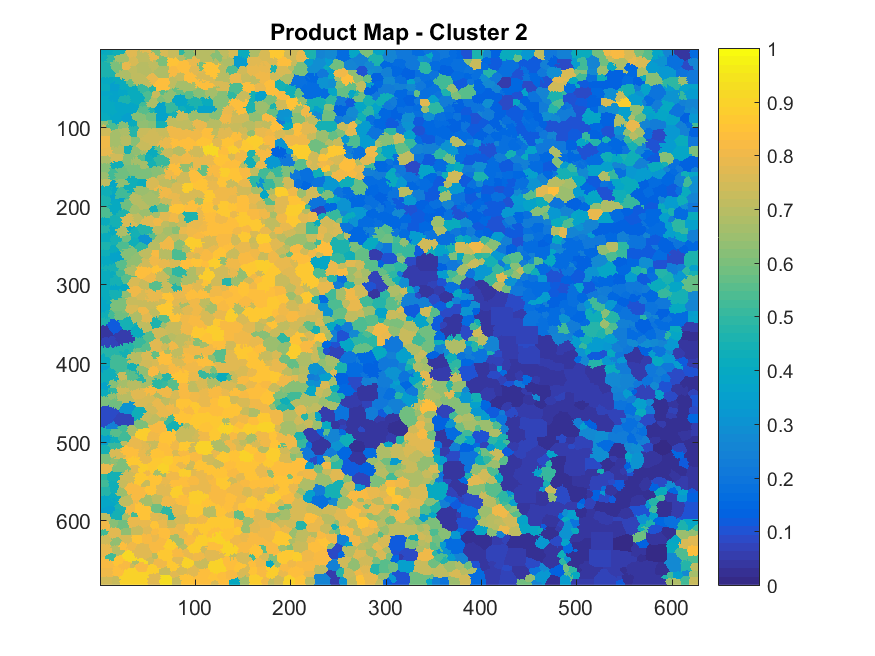}}}
		\subfigure[ FLICM Cluster 3 ]{\resizebox{!}{.95in}{\includegraphics[trim={1.6cm 1cm 3cm .75cm},clip]{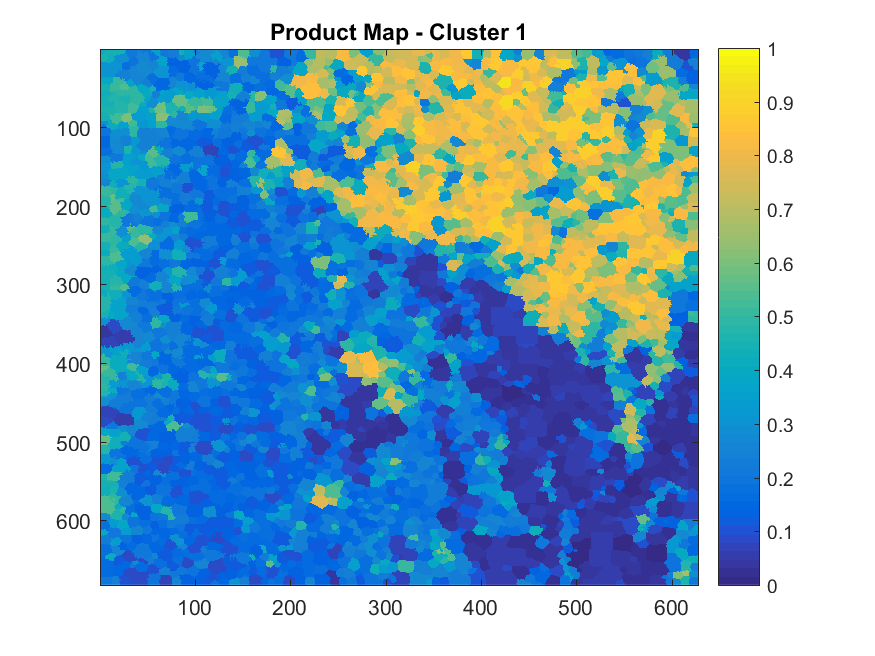}}}\\
		\subfigure[ PFCM Cluster 1 ]{\resizebox{!}{.95in}{\includegraphics[trim={1.6cm 1cm 3cm .75cm},clip]{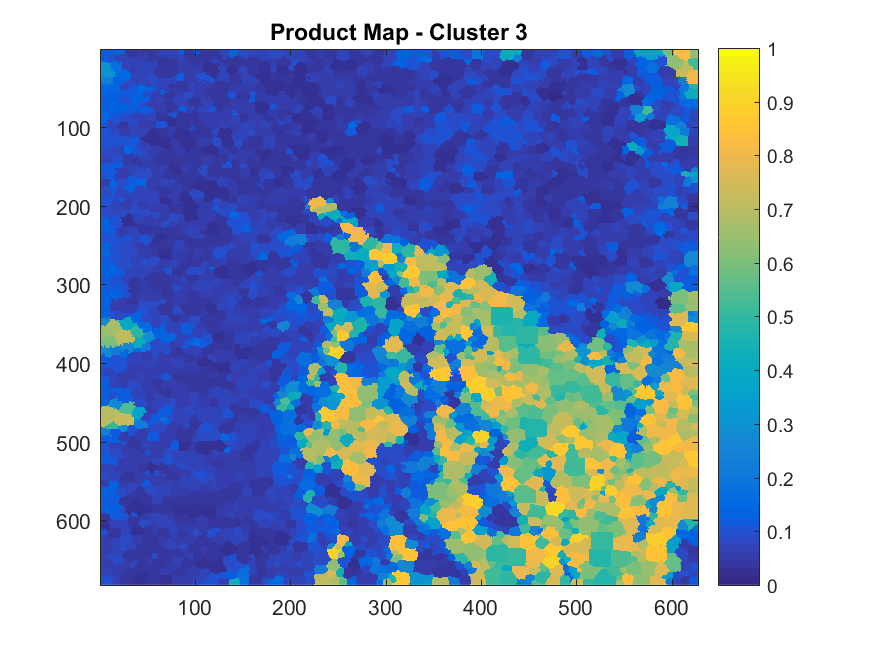}}}
		\subfigure[ PFCM Cluster 2 ]{\resizebox{!}{.95in}{\includegraphics[trim={1.6cm 1cm 3cm .75cm},clip]{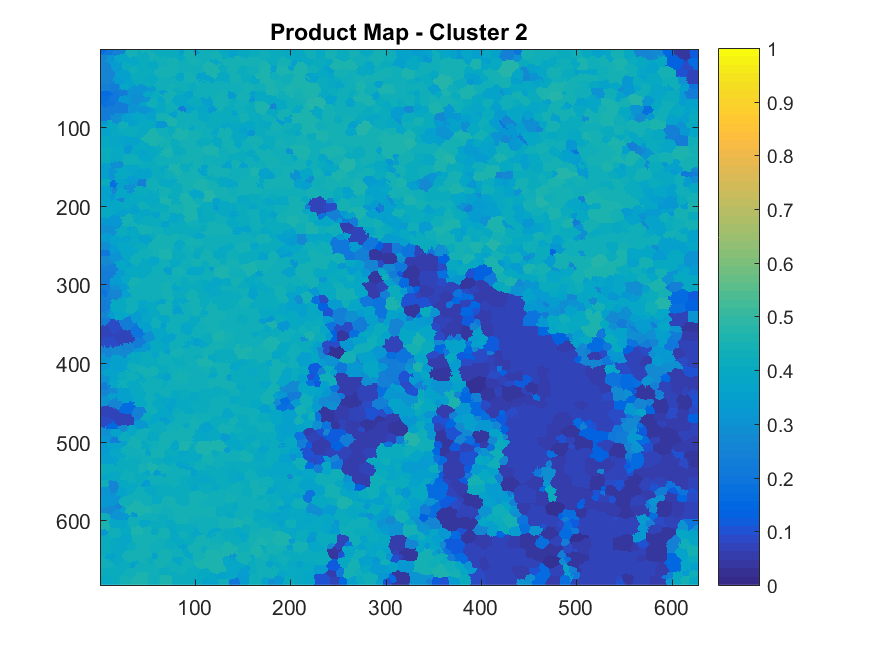}}}
		\subfigure[ PFCM Cluster 3 ]{\resizebox{!}{.95in}{\includegraphics[trim={1.6cm 1cm 3cm .75cm},clip]{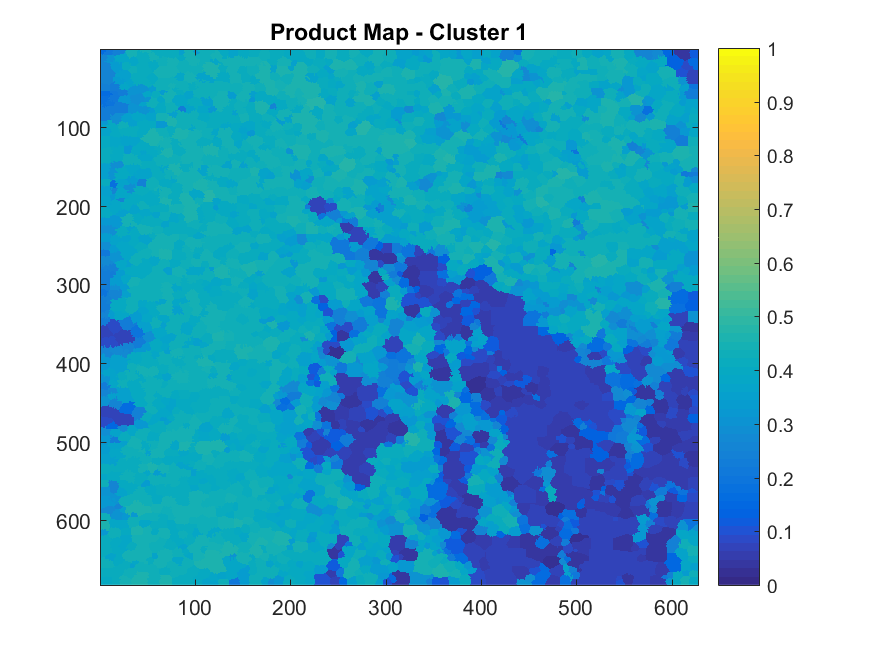}}}\\
     		\caption[]{ (a) Image containing sand ripples, smooth sand, sand with hilly regions and shadows. Clustering results of image (a) given by the PFLICM (b-d), FLICM (e-g), and PFCM (h-j) algorithms. Clusters have been manually aligned for easy comparison. }
		\label{fig:im5}		}
\end{figure}

\begin{figure}[!t]
	\center{
			\subfigure[ Original Image ]{\resizebox{!}{2.5in}{\includegraphics[trim={1cm 0cm 1cm .75cm},clip]{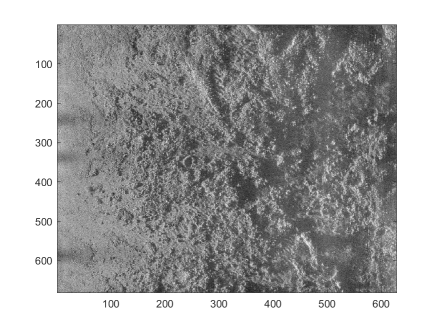}}}\\
		\subfigure[ PFLICM Cluster 1 ]{\resizebox{!}{.95in}{\includegraphics[trim={1.6cm 1cm 3cm .75cm},clip]{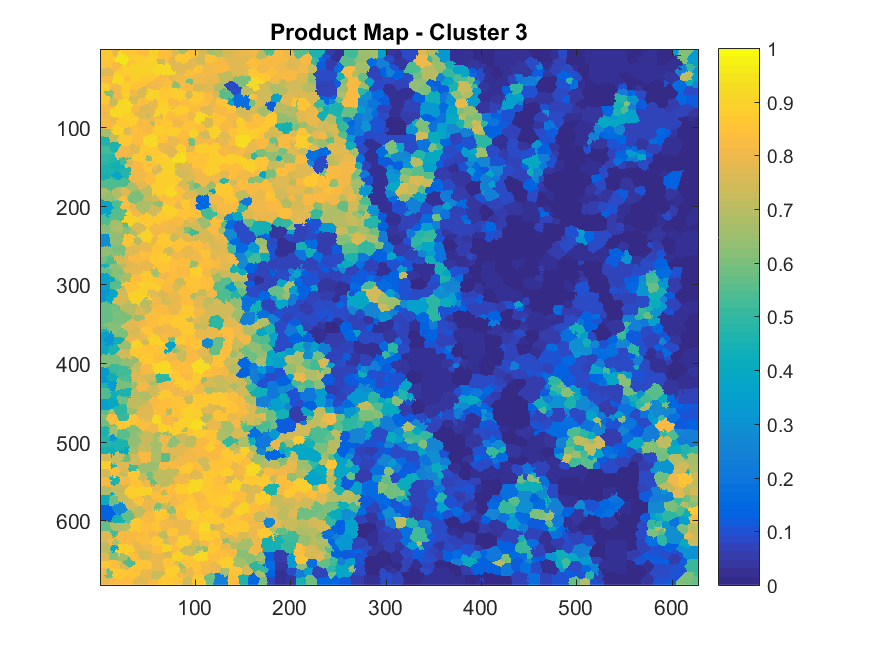}}}
		\subfigure[ PFLICM Cluster 2 ]{\resizebox{!}{.95in}{\includegraphics[trim={1.6cm 1cm 3cm .75cm},clip]{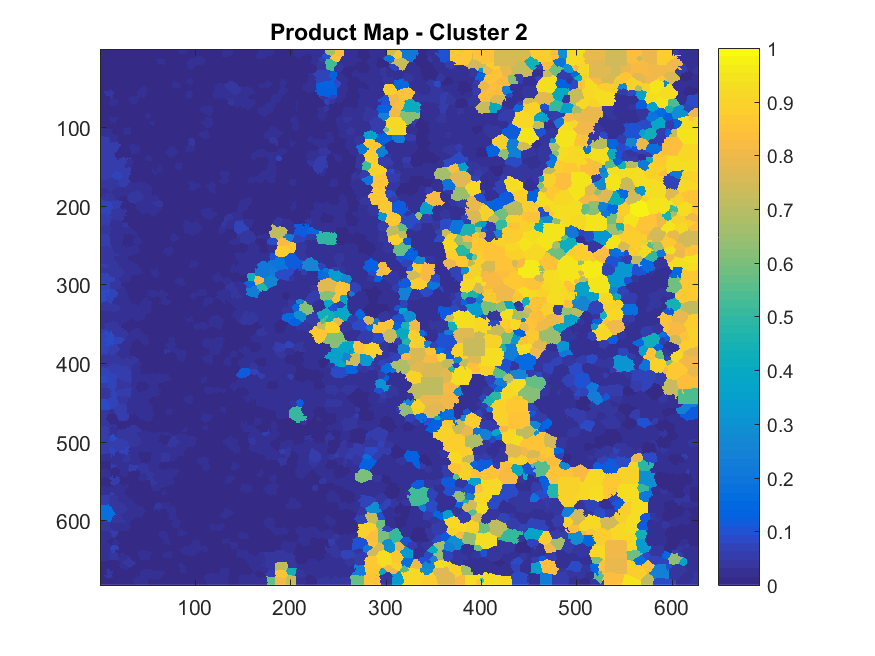}}}
		\subfigure[ PFLICM Cluster 3 ]{\resizebox{!}{.95in}{\includegraphics[trim={1.6cm 1cm 3cm .75cm},clip]{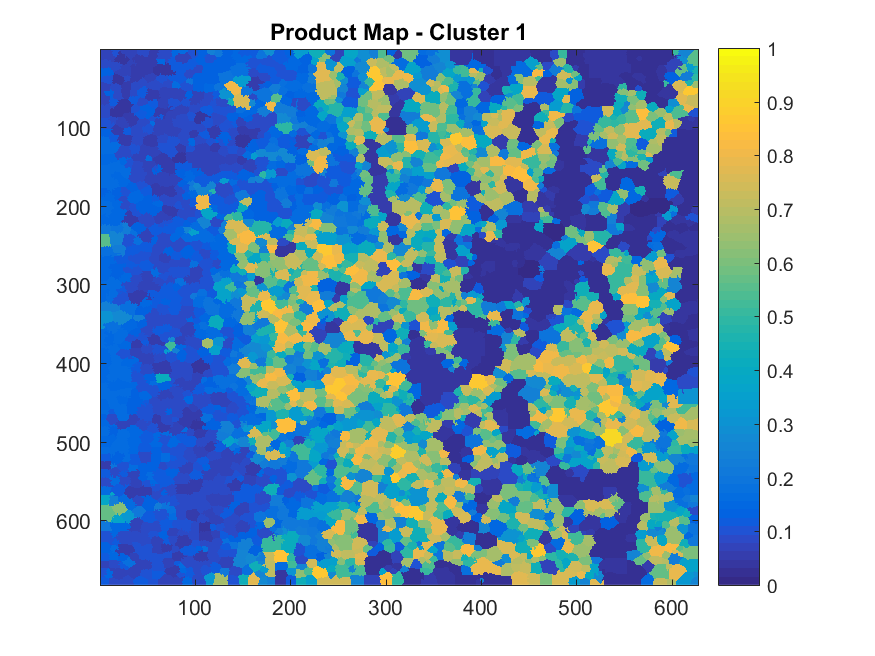}}}\\
		\subfigure[ FLICM Cluster 1 ]{\resizebox{!}{.95in}{\includegraphics[trim={1.6cm 1cm 3cm .75cm},clip]{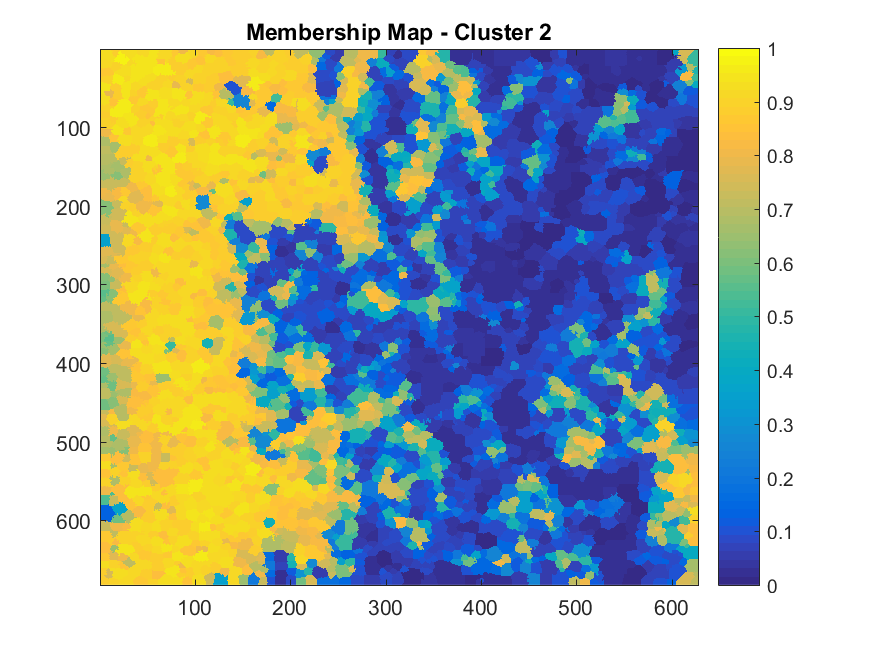}}}
		\subfigure[ FLICM Cluster 2 ]{\resizebox{!}{.95in}{\includegraphics[trim={1.6cm 1cm 3cm .75cm},clip]{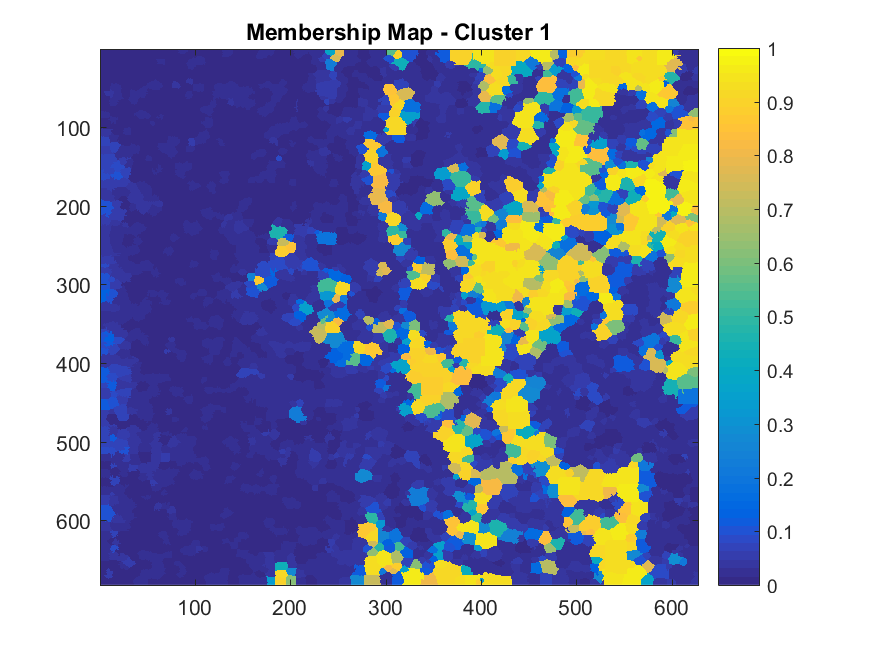}}}
		\subfigure[ FLICM Cluster 3 ]{\resizebox{!}{.95in}{\includegraphics[trim={1.6cm 1cm 3cm .75cm},clip]{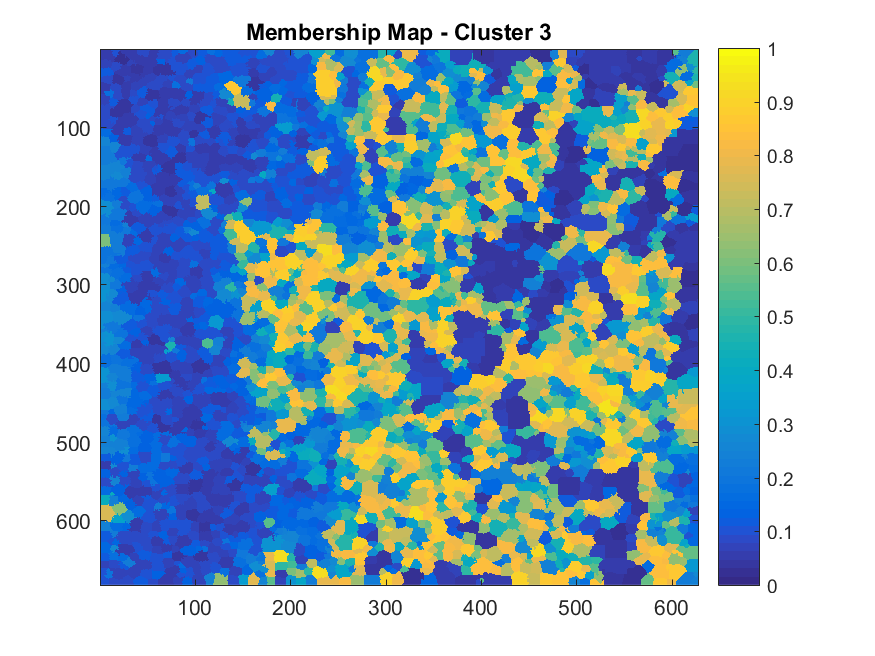}}}\\
		\subfigure[ PFCM Cluster 1 ]{\resizebox{!}{.95in}{\includegraphics[trim={1.6cm 1cm 3cm .75cm},clip]{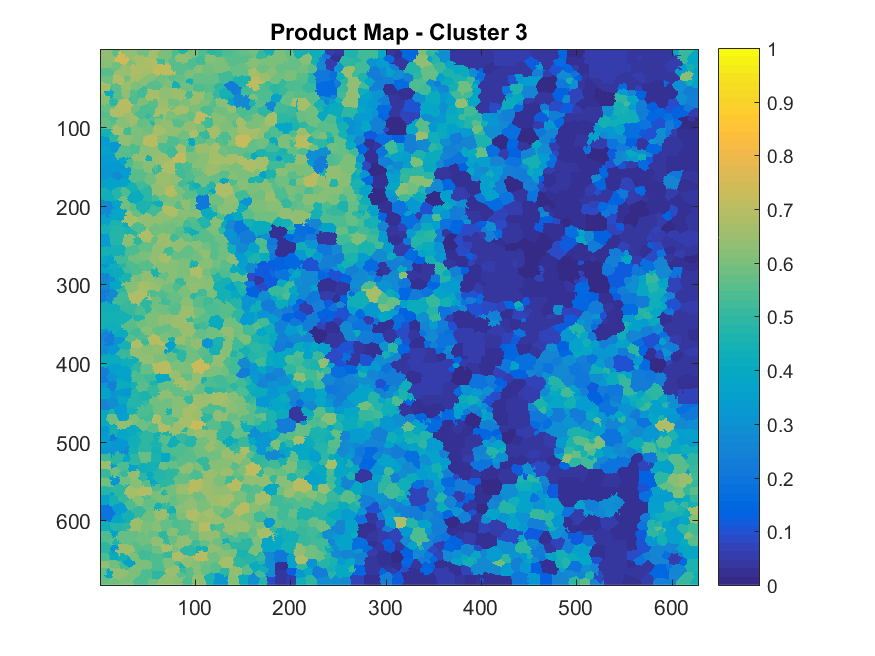}}}
		\subfigure[ PFCM Cluster 2 ]{\resizebox{!}{.95in}{\includegraphics[trim={1.6cm 1cm 3cm .75cm},clip]{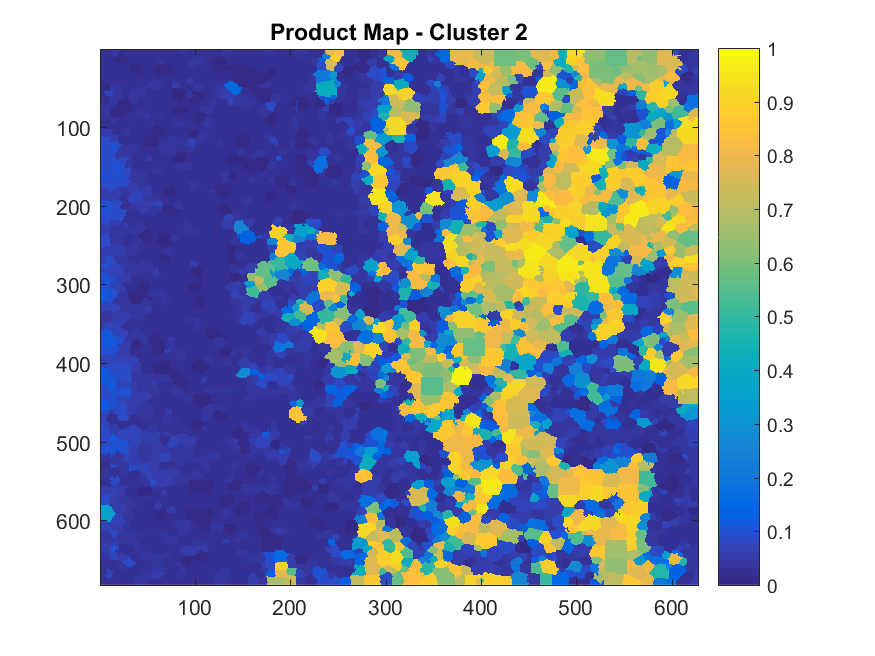}}}
		\subfigure[ PFCM Cluster 3 ]{\resizebox{!}{.95in}{\includegraphics[trim={1.6cm 1cm 3cm .75cm},clip]{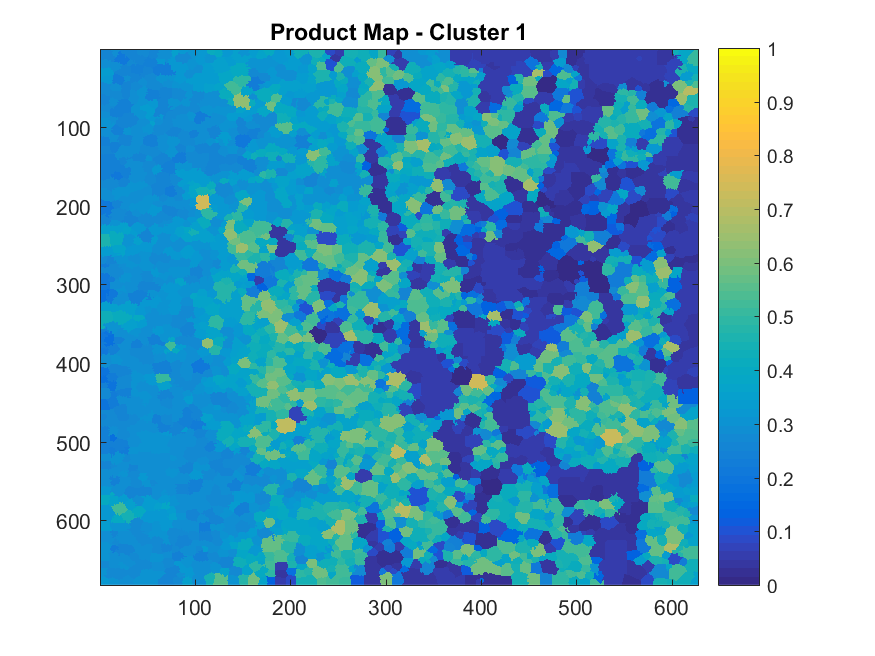}}}\\
     		\caption[]{ (a) Image containing smooth sand and a great variety of hilly regions with shadows. Clustering results of image (a) given by the PFLICM (b-d), FLICM (e-g), and PFCM (h-j) algorithms. Clusters have been manually aligned for easy comparison. }
		\label{fig:im6}		}
\end{figure}

\begin{figure}[!t]
	\center{
		\subfigure[ Original Image ]{\resizebox{!}{2.5in}{\includegraphics[trim={1cm 0cm 1cm .75cm},clip]{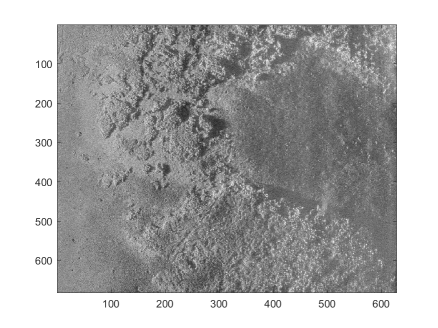}}}\\
		\subfigure[ PFLICM Cluster 1 ]{\resizebox{!}{.95in}{\includegraphics[trim={1.6cm 1cm 3cm .75cm},clip]{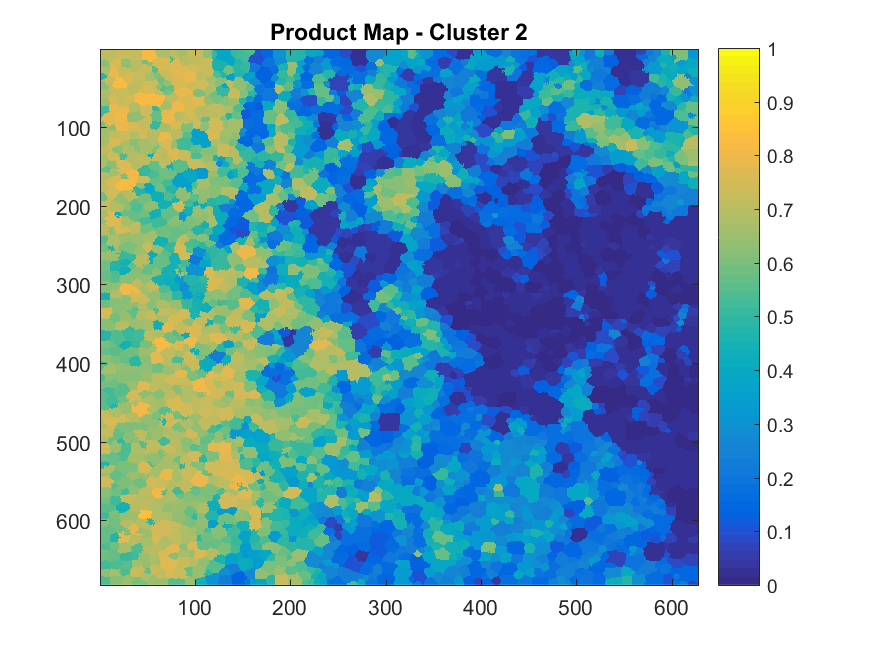}}}
		\subfigure[ PFLICM Cluster 2 ]{\resizebox{!}{.95in}{\includegraphics[trim={1.6cm 1cm 3cm .75cm},clip]{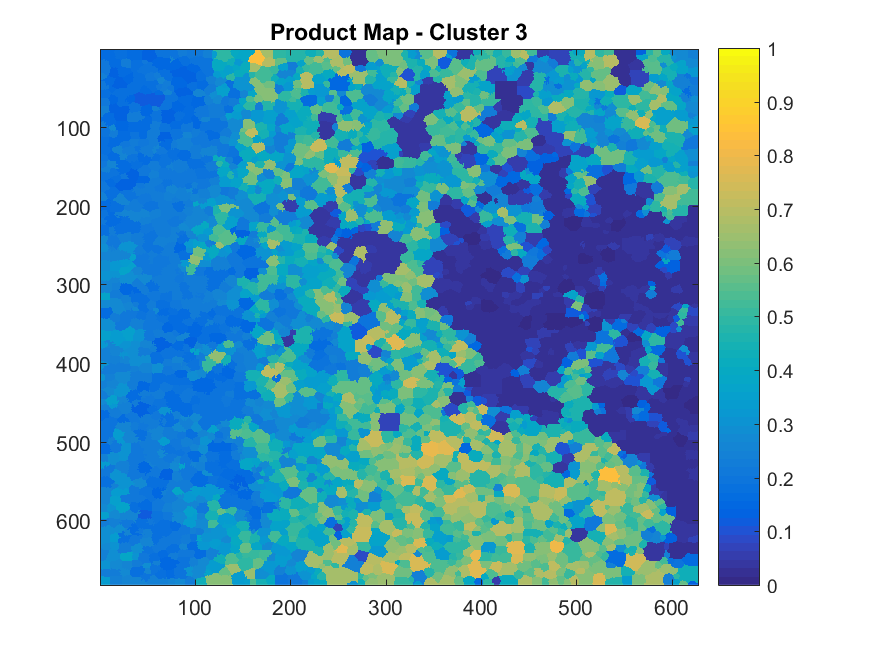}}}
		\subfigure[ PFLICM Cluster 3 ]{\resizebox{!}{.95in}{\includegraphics[trim={1.6cm 1cm 3cm .75cm},clip]{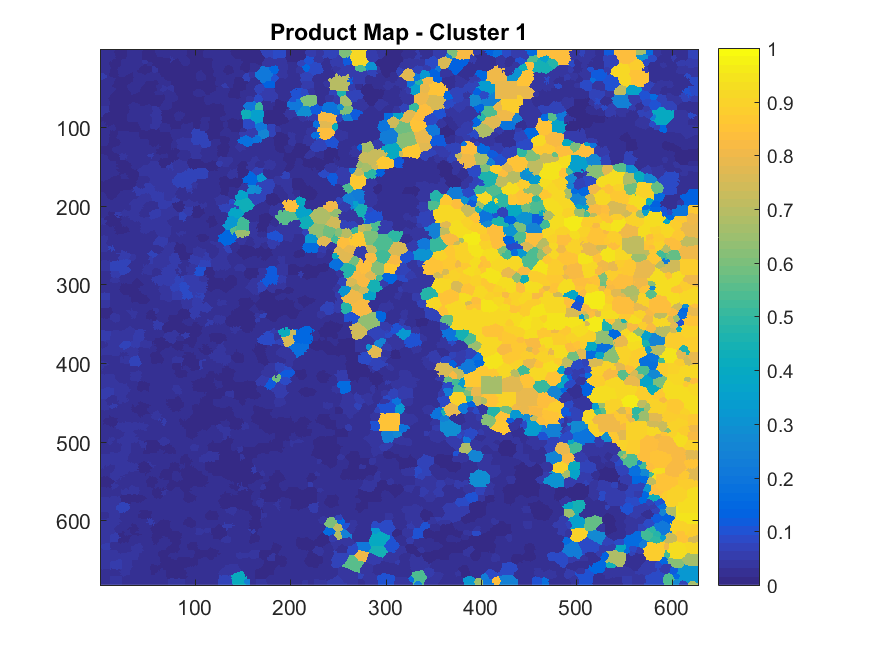}}}\\
		\subfigure[ FLICM Cluster 1 ]{\resizebox{!}{.95in}{\includegraphics[trim={1.6cm 1cm 3cm .75cm},clip]{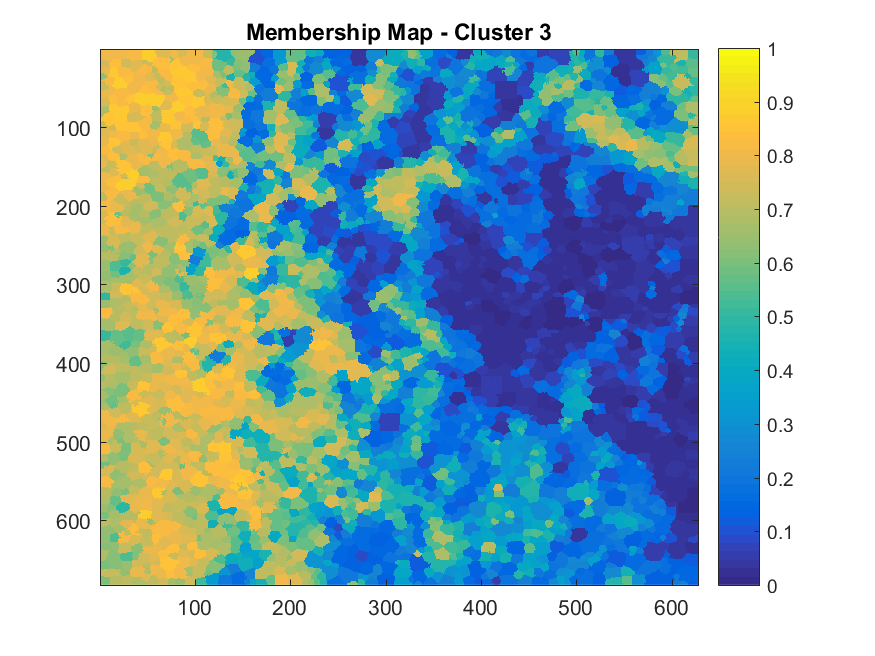}}}
		\subfigure[ FLICM Cluster 2 ]{\resizebox{!}{.95in}{\includegraphics[trim={1.6cm 1cm 3cm .75cm},clip]{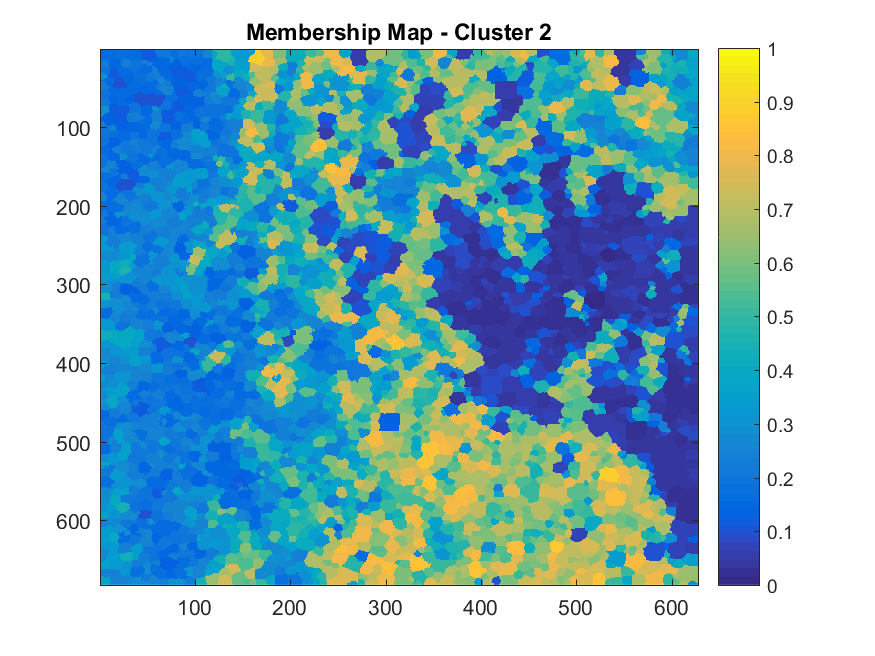}}}
		\subfigure[ FLICM Cluster 3 ]{\resizebox{!}{.95in}{\includegraphics[trim={1.6cm 1cm 3cm .75cm},clip]{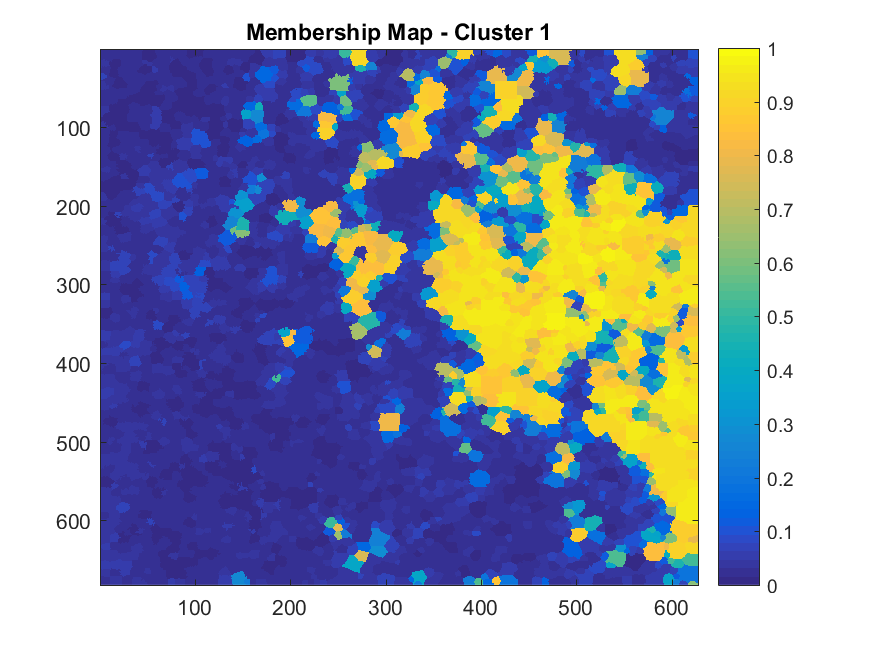}}}\\
		\subfigure[ PFCM Cluster 1 ]{\resizebox{!}{.95in}{\includegraphics[trim={1.6cm 1cm 3cm .75cm},clip]{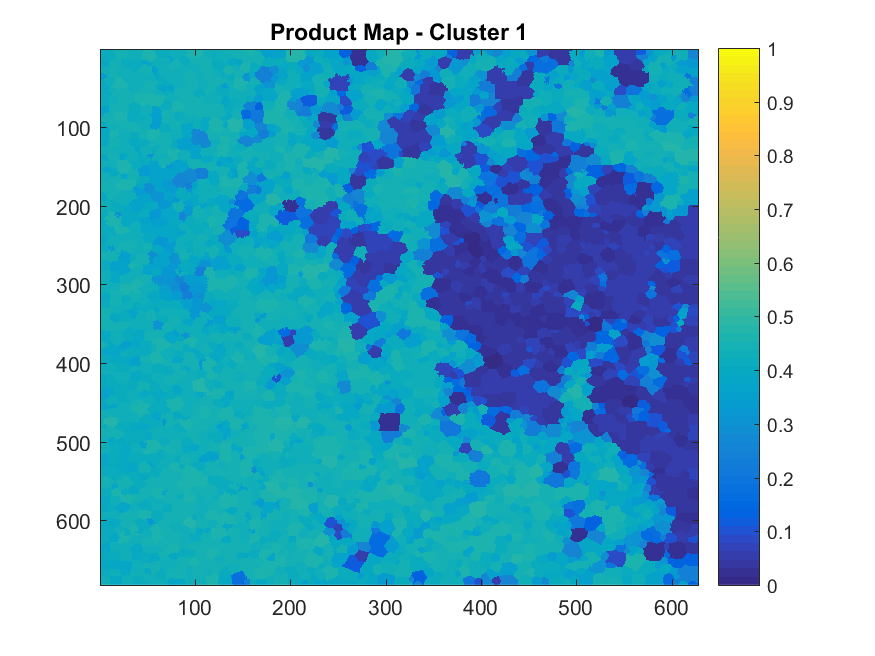}}}
		\subfigure[ PFCM Cluster 2 ]{\resizebox{!}{.95in}{\includegraphics[trim={1.6cm 1cm 3cm .75cm},clip]{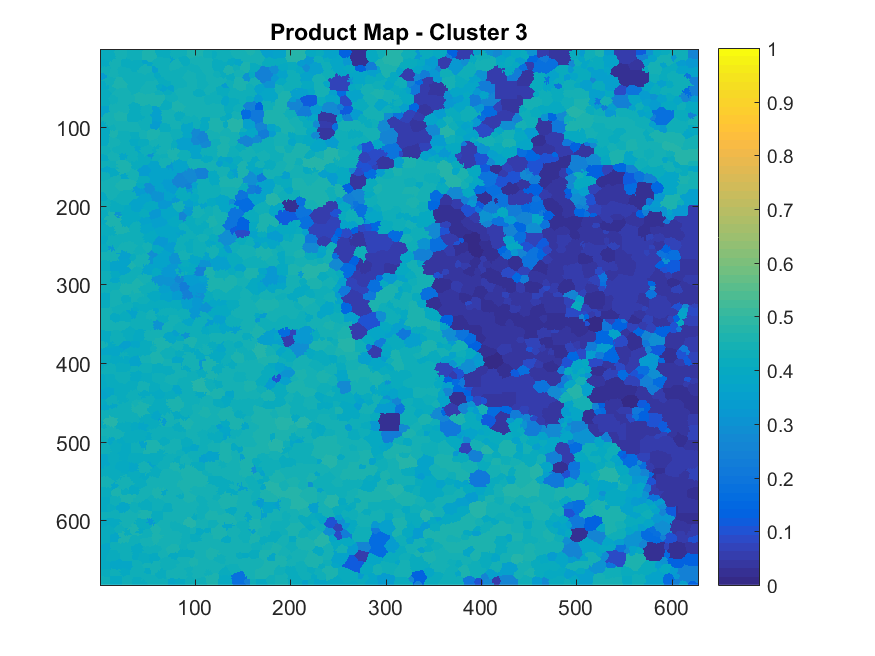}}}
		\subfigure[ PFCM Cluster 3 ]{\resizebox{!}{.95in}{\includegraphics[trim={1.6cm 1cm 3cm .75cm},clip]{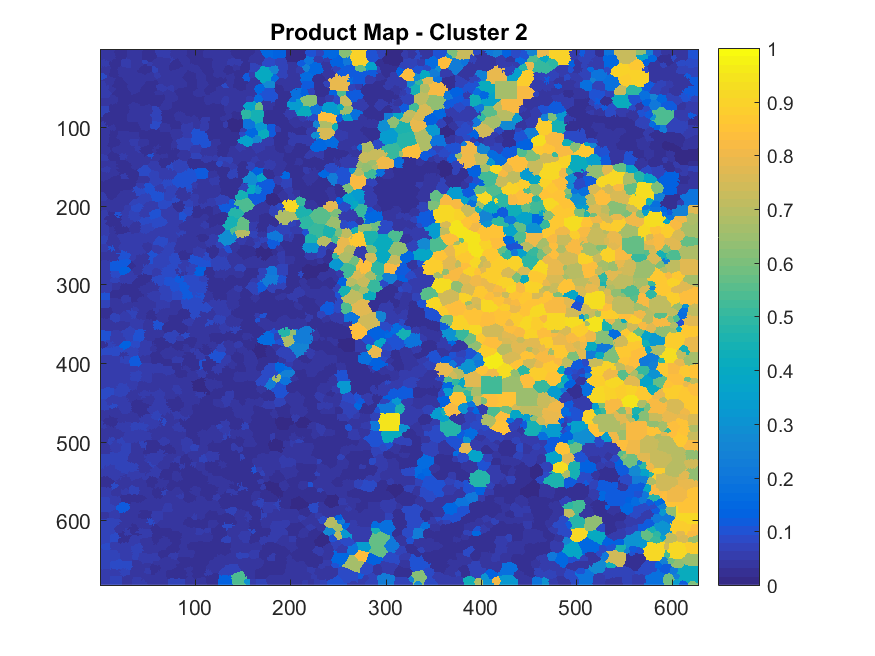}}}\\
     		\caption[]{(a) Image containing smooth sand, hard-packed sand, and hilly regions with shadows. Clustering results of image (a) given by the PFLICM (b-d), FLICM (e-g), and PFCM (h-j) algorithms. Clusters have been manually aligned for easy comparison. }
		\label{fig:im7}		}
\end{figure}

Since PFLICM  estimates typicality in addition to membership, it is able to disregard potential outliers. Figure \ref{fig:im4a} shows the typicality maps of PFLICM and PFCM and the membership map of FLICM. Note that the typicality maps of PFLICM and PFCM are identical because the fuzzy factor $G_{cn}$ does not contribute to typicality. Figure \ref{fig:im4b} provides zoomed in images on the typicality and membership maps of Figure \ref{fig:im4a} (Typicality maps only shown once to avoid redundancy because of the aforementioned equality). Notice how the typicality maps (b-d) clearly diminish the value at the outlier, while the membership maps (e-g) do not. Thus, PFLICM maintains the ability of PFCM to identify and effectively nullify the contribution of outliers.

\clearpage
Figures 5 - 10 provide additional evidence of the utility of this spatial fuzzy and possibilistic approach to seafloor segmentation.  
These results illustrate how PFLICM blends the
strengths of both FLICM and PFCM into a single set of results.
Namely, the PFLICM algorithm maintains the membership
smoothness and encourages sparse membership values when
appropriate (as found with FLICM) while still mitigating the
effects of outliers (an attribute characteristic of PFCM).

 \clearpage

\section{Summary}
The possibilistic fuzzy local information c-means clustering algorithm is developed and presented with application to SAS image segmentation.  PFLICM is shown to combine the advantages of multiple alternative segmentation approaches. Results show improvement over existing soft segmentation approaches and successfully identify various sea floor textures comparable to the ability of a human observer.

% conference papers do not normally have an appendix

% use section* for acknowledgment
\section*{Acknowledgment}
This research was funded by the Office of Naval Research Code 321.

% trigger a \newpage just before the given reference
% number - used to balance the columns on the last page
% adjust value as needed - may need to be readjusted if
% the document is modified later
%\IEEEtriggeratref{8}
% The "triggered" command can be changed if desired:
%\IEEEtriggercmd{\enlargethispage{-5in}}

% references section

% can use a bibliography generated by BibTeX as a .bbl file
% BibTeX documentation can be easily obtained at:
% http://mirror.ctan.org/biblio/bibtex/contrib/doc/
% The IEEEtran BibTeX style support page is at:
% http://www.michaelshell.org/tex/ieeetran/bibtex/
%\bibliographystyle{IEEEtran}
% argument is your BibTeX string definitions and bibliography database(s)
%\bibliography{IEEEabrv,../bib/paper}
%
% <OR> manually copy in the resultant .bbl file
% set second argument of \begin to the number of references
% (used to reserve space for the reference number labels box)
\bibliographystyle{IEEEtran}
\bibliography{Reference}

% that's all folks
\end{document}